\documentclass[11pt]{article}

\usepackage[final]{acl}
\usepackage{amsmath, bm} % 加载常用数学包
\usepackage{times}
\usepackage{latexsym}

\usepackage[T1]{fontenc}
\usepackage[utf8]{inputenc}

\usepackage{microtype}

\usepackage{inconsolata}

\usepackage{graphicx}

\usepackage{booktabs} % 三线表必备宏包
\usepackage{multirow} % 合并单元格
\usepackage{hyperref}
\usepackage{amsfonts}

\usepackage{amsmath}
\usepackage{amssymb}
\usepackage{xcolor}
\usepackage{tcolorbox}
\usepackage{listings}

\usepackage{tabularx}
\usepackage{array}
\usepackage{bm}
\usepackage{pifont}

\definecolor{TCEBackground}{RGB}{235, 242, 250}

\newcommand{\cmark}{\ding{51}}
\newcommand{\xmark}{\ding{55}}
\newcommand{\pmark}{$\triangle$}

\tcbset{tceboxstyle/.style={
    colback=TCEBackground, 
    colframe=TCEBackground, 
    boxrule=0pt, 
    arc=3pt, 
    boxsep=3pt,      % 稍微留一点呼吸感 (原为5pt)
    top=0pt,         % 去除顶部额外留白
    bottom=0pt,      % 去除底部额外留白
    left=2pt, right=2pt,
    standard jigsaw
}}

\title{TangramPuzzle: Evaluating Multimodal Large Language Models with Compositional Spatial Reasoning}

\author{
Daixian Liu$^{1}$\thanks{\ \ indicates equal contribution.},~Jiayi Kuang$^{3*}$,~Yinghui Li$^{4}$,~Yangning Li$^{1}$,~\textbf{Di Yin}$^{4}$,~\textbf{Haoyu Cao}$^{4}$\\
~\textbf{Xing Sun}$^{4}$,~\textbf{Ying Shen}$^{3}$,\textbf{Hai-Tao Zheng}$^{1,2}$%
\thanks{\mbox{\footnotesize Corresponding author: \footnotesize{zheng.haitao@sz.tsinghua.edu.cn}.}},~\textbf{Liang Lin}$^{3}$,~\textbf{Philip S. Yu}$^{5}$\\
$^{1}$ Shenzhen International Graduate School, Tsinghua University \\
$^{2}$ Pengcheng Laboratory, Shenzhen, China,
$^{3}$ Sun Yat-sen University \\
$^{4}$ Tencent Youtu Lab, $^{5}$ University of Illinois Chicago \\
\texttt{liudx25@mails.tsinghua.edu.cn}, \texttt{liyinghuihhh@gmail.com}
}

\begin{document}
\maketitle
\begin{abstract}
Multimodal Large Language Models (MLLMs) have achieved remarkable progress in visual recognition and semantic understanding, yet precise compositional spatial reasoning under geometric constraints remains underexplored. Existing benchmarks mainly assess coarse spatial relations and rarely support rigorous geometric verification or multiple valid solutions in constructive tasks. To address these limitations, we introduce \textbf{TangramPuzzle}~\footnote{https://github.com/THUKElab/TangramPuzzle}, a benchmark comprising 668 validated configurations and 1,336 instances for evaluating compositional spatial reasoning under geometric constraints. We formulate the Tangram Construction Expression (TCE) to encode tangram configurations with machine-verifiable geometric specifications, together with a multi-solution-aware verifier that separately evaluates geometric validity and silhouette fidelity. TangramPuzzle covers two complementary evaluation tasks: part-to-whole \emph{Outline Prediction} and whole-to-part \emph{End-to-End Assembly Generation}. We conduct extensive evaluation experiments on advanced open-source and proprietary models, revealing an interesting insight: MLLMs tend to prioritize matching the target silhouette while neglecting geometric constraints, leading to distortions or deformations of the pieces.

\end{abstract}

\section{Introduction}
Spatial reasoning has become an increasingly important capability for Multimodal Large Language Models (MLLMs). Early models primarily focused on basic spatial perception tasks, such as relative distance estimation and object orientation recognition~\citep{wang2025spatial457, fang2025guided,ma2025deepperception}. More recent research has shifted toward finer-grained and more rigorous spatial reasoning, as exemplified by visual grounding~\citep{tang2025visual, xu2025mc}, Game agents~\citep{ouyang2026gameworld,cheng2025kairos}, and visual reasoning in complex scenes~\citep{kuang2025express}, which demand precise localization, interaction, and reasoning over complex spatial structures. Robust spatial reasoning therefore requires constructing and validating geometric relationships beyond relative positioning. Robust spatial reasoning therefore goes beyond identifying relative positions; it requires constructing and validating structured geometric relationships for reliable decision making.

\begin{figure}
    \centering
    \includegraphics[width=\linewidth]{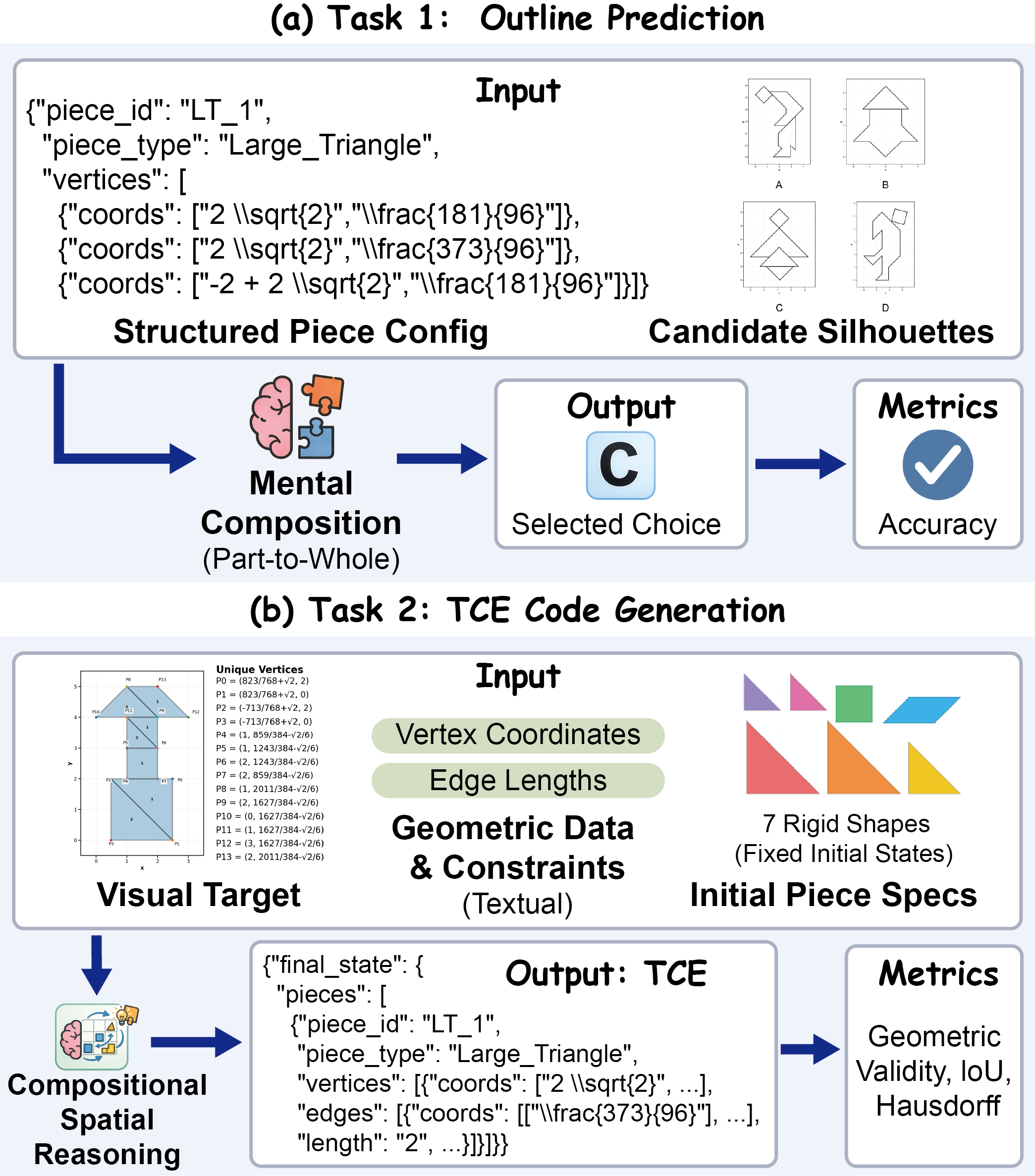}
    % \caption{Illustration of the two spatial reasoning tasks. (a) Task 1 global shape inference; (b) Task 2 constrained shape decomposition.}
    \caption{Illustration of the two spatial reasoning tasks: (a) Task 1 for global shape inference from piece configurations; (b) Task 2 for constrained shape decomposition.}
    \label{fig:introduction}
\end{figure}

Despite rapid progress, existing spatial reasoning tasks for MLLMs still face critical limitations. First, many tasks focus on \textbf{coarse relational concepts} (e.g., left/right, front/behind, adjacency)~\citep{liu2025can}, which do not fully capture reasoning under strict geometric and physical constraints. Second, spatial configurations are often specified through \textbf{natural-language descriptions} or multiple-choice questions rather than formal geometric representations~\citep{ma20253dsrbench}, making them \textbf{ambiguous and difficult to verify} precisely. Third, evaluation commonly relies on answer matching, with limited consideration of the \textbf{multiple valid solutions} inherent in constructive tasks~\citep{rodionov2025floorplanqa}. In particular, similarity-based metrics may reward geometrically invalid assemblies: high IoU can result from stretched or overlapping pieces. Thus, we wonder \textbf{\textit{how should compositional spatial reasoning be evaluated when correctness requires both target fidelity and part-level validity?}}

Inspired by the classic Tangram puzzle~\citep{bofferding2023does}, we study \textbf{compositional spatial reasoning under strict geometric constraints}. Solving a tangram requires arranging seven polygonal pieces through translation, rotation, and optional reflection to exactly fill a target silhouette, while preserving piece rigidity, avoiding overlap, and maintaining topological validity. These operations abstract key capabilities underlying practical problems such as CAD-like assembly, robotic manipulation, and structural reconstruction. Yet existing tangram resources primarily emphasize semantic interpretation or perceptual matching~\citep{zhang2024tangram, ji2022abstract}, leaving exact, machine-verifiable assembly reasoning largely unaddressed.

To address these gaps, we introduce \textbf{TangramPuzzle}, a benchmark for evaluating compositional spatial reasoning in MLLMs under strict geometric constraints. We define the \textbf{Tangram Construction Expression (TCE)}, a symbolic geometry schema that represents each tangram configuration with exact coordinates and structural relations, replacing ambiguous prose with machine-verifiable specifications. Building on TCE, TangramPuzzle evaluates two directions of spatial reasoning, as shown in Figure~\ref{fig:introduction}: (1) \textbf{\textit{part-to-whole composition}} through \emph{Outline Prediction}, where models infer a global silhouette from locally specified pieces; and (2) \textbf{\textit{whole-to-part decomposition}} through \emph{End-to-End Assembly Generation}, where models decompose a target outline into a complete configuration of seven rigid pieces. A multi-solution-aware verifier evaluates syntactic validity, rigid-shape preservation, physical feasibility, and silhouette fidelity using IoU and Hausdorff distance, without relying on a single reference assembly.

TangramPuzzle is constructed through a multi-stage pipeline comprising source filtering, interactive annotation with geometric snapping, symbolic normalization into exact expressions, and human-in-the-loop validation. Extensive evaluations across diverse MLLMs show that the benchmark remains highly challenging: \textbf{\textit{models often approximate the target silhouette while violating piece rigidity and physical constraints,}} revealing systematic failures in both part-to-whole composition and constraint-satisfying whole-to-part decomposition. Our main contributions are as follows:
\begin{itemize}
    % \item We introduce TangramPuzzle, a benchmark comprising 668 validated configurations and 1,336 evaluation instances across two tasks: forward composition from local pieces to a global shape and inverse decomposition from a target shape to a valid assembly.

    \item We introduce TangramPuzzle, a benchmark comprising 668 validated configurations and 1,336 evaluation instances across two tasks: part-to-whole composition from local pieces to a global shape and whole-to-part decomposition from a target shape to a valid assembly.

    % \item We propose the Tangram Construction Expression (TCE) and a multi-solution-aware verifier to enable exact, machine-verifiable representation and precise evaluation of tangram assemblies.

    \item We propose the Tangram Construction Expression (TCE) and a multi-solution-aware verifier for exact, machine-verifiable representation and fine-grained evaluation of compositional spatial reasoning under geometric constraints, covering rigid-body preservation, physical feasibility, and silhouette fidelity.
    
    \item Our evaluation reveals a substantial gap between fidelity and validity in MLLMs, which often match target silhouettes by violating rigidity or non-overlap constraints.
\end{itemize}

\section{Related Work}
\subsection{General Multimodal Benchmarks}
Standard benchmarks for Multimodal Large Language Models (MLLMs) have transitioned from foundational tasks like visual question answering (VQA)~\cite{cheng2025simplevqa, zhong2025focus} and image captioning~\cite{dong2024benchmarking,lu2025benchmarking} to comprehensive assessments of emergent capabilities. Recent benchmarks target diverse competencies, including reasoning ability~\cite{yuan2025mme, xiao2024logicvista}, where challenging datasets like Humanity’s Last Exam~\cite{phan2025humanity} and ZeroBench~\cite{roberts2025zerobench} test the upper bounds of expert-level cognition; document and chart understanding~\cite{wang2024charxiv,xia2025chartx, ouyang2025omnidocbench}; as well as visual grounding~\cite{paiss2023teaching, brazil2023omni3d}, hallucination evaluation~\cite{guan2024hallusionbench, yang2025wildvideo}, and multi-image comprehension~\cite{cheng2025evaluating, fu2024blink}. Furthermore, the assessment of agentic capabilities~\cite{li2025screenspot, xie2025scaling} centers on GUI interpretation and precise element interaction.

\subsection{Spatial Reasoning Benchmarks}
\begin{table}[t]
\centering
\scriptsize
\renewcommand{\arraystretch}{1.16}
\setlength{\tabcolsep}{1.8pt}
\caption{
Comparison with existing spatial reasoning benchmarks.
InvGen: structured inverse generation;
RigidRep: explicit rigid-part representation;
ProgVer: programmatic constraint verification;
MultiSol: multi-solution-aware evaluation.
\cmark, \xmark, and \pmark indicate full, absent, and partial support.
}
\begin{tabular*}{\columnwidth}{
    @{\extracolsep{\fill}}
    l l c c c c
    @{}
}
\toprule
\textbf{Benchmark}
& \textbf{Output}
& \textbf{InvGen}
& \textbf{RigidRep}
& \textbf{ProgVer}
& \textbf{MultiSol}
\\
\midrule

OmniSpatial
& VQA
& \xmark & \xmark & \xmark & \xmark
\\

MMSI-Bench
& MCQ
& \xmark & \xmark & \xmark & \xmark
\\

LEGO-Puzzles
& Plan/Img.
& \pmark & \xmark & \xmark & \xmark
\\

Tangram
& Counting
& \xmark & \xmark & \xmark & \xmark
\\

ORIGAMISPACE
& CP Code
& \cmark & \xmark & \cmark & \xmark
\\

\midrule

\textbf{TangramPuzzle}
& \textbf{TCE}
& \cmark & \cmark & \cmark & \cmark
\\

\bottomrule
\end{tabular*}
\label{tab:benchmark_comparison}
\end{table}

Spatial reasoning is essential for MLLMs to perceive and interact with the physical world~\cite{zheng2025multimodal}. Early benchmarks, such as SpatialBench~\cite{xu2025spatialbench} and RealWorldQA~\cite{xai2024realworldqa}, mainly assess relative position and depth, while OmniSpatial~\cite{jia2025omnispatial}, MMSI-Bench~\cite{yang2025mmsi}, and LEGO-Puzzles~\cite{tang2025lego} extend evaluation to dynamic scenes, multi-view reasoning, and sequential assembly. However, most remain VQA-based and may encourage shortcuts based on language priors or superficial visual cues. ORIGAMISPACE~\cite{xuorigamispace} evaluates procedural folding, whereas TangramPuzzle targets inverse assembly of multiple rigid components under exact geometric and physical constraints. In the tangram domain, Tangram~\cite{zhang2024tangram} focuses on element recognition and counting, while TANGAN~\cite{yamada2025tangan} treats assembly as an optimization problem rather than an MLLM benchmark. As shown in Table~\ref{tab:benchmark_comparison}, TangramPuzzle uniquely combines structured inverse assembly, exact rigid-body validation, and multi-solution-aware evaluation. Further comparison is provided in Appendix~\ref{app:benchmark_comparison}.

% Unlike prior approaches that rely mainly on perceptual approximations, our benchmark is built around explicit mathematical representations and constraint-based evaluation, enabling precise assessment of solution validity. Its expressive 2D formulation also avoids the viewpoint and depth ambiguities common in 3D tasks, such as self-occlusion and projection effects, which often make errors difficult to attribute clearly to reasoning failures. In contrast, the 2D formulation makes reasoning failures clearer and supports automatic verification.

% % Its expressive 2D formulation also avoids the viewpoint and depth ambiguities inherent in 3D tasks, making reasoning failures clearer and automatically verifiable.

\section{Method}
\subsection{Task Definition}
% The Tangram is a classic dissection puzzle requiring the arrangement of seven rigid polygons to form specific shapes, serving as an ideal testbed for compositional spatial reasoning. While natural language is expressive, relying solely on verbal descriptions for geometric assembly leads to substantial ambiguity and imprecision, making reliable evaluation difficult due to the lack of spatial granularity. In contrast, TangramPuzzle is governed by strict Euclidean geometry and topological constraints, motivating a formal, coordinate-based representation that grounds visual inputs in precise mathematical formulations. This design enables unambiguous task definition, automatic verification, and fine-grained evaluation of spatial reasoning abilities. TangramPuzzle consists of two challenging tasks designed to evaluate the spatial reasoning capabilities of MLLMs, ranging from passive spatial perception to active solution generation.

Tangram is a classic dissection puzzle in which seven rigid polygons must be assembled to form a target shape. Although simple in form, tangram assembly mirrors practical tasks such as assembling parts into a target structure, where each component must be placed with the correct orientation and position for the overall configuration to be valid, making it an ideal testbed for compositional spatial reasoning. TangramPuzzle consists of two challenging tasks designed to evaluate the spatial reasoning capabilities of MLLMs, ranging from passive spatial perception to active solution generation.

\paragraph{Geometry Description by TCE Formula}

% While natural language is expressive, verbal descriptions alone are often too ambiguous for geometric assembly. For example, ``place the triangle above the square'' may refer either to a triangle directly touching the square from above or to one located some distance above it, making reliable evaluation difficult because multiple geometrically distinct configurations may satisfy the same description. 

Natural language descriptions are often ambiguous for geometric assembly, where a phrase like ``place the triangle above the square'' fails to specify whether the pieces touch or separate. Moreover, TangramPuzzle is governed by strict Euclidean and topological constraints, motivating a formal, coordinate-based representation that grounds visual inputs in precise mathematical formulations and supports unambiguous task definition, automatic verification, and fine-grained evaluation of spatial reasoning abilities. Hence, the Tangram Construction Expression (TCE) is introduced as a symbolic geometry description language for 2D tangram assembly. Each TCE instance consists of five core components: instance\_id, target\_outline, initial\_state, final\_state, and adjacency\_graph. The representation of TCE is shown in Figure~\ref{fig:tce_definition}. Detailed TCE definitions can be found in Appendix~\ref{app:tce_definition}.

\begin{figure}[htbp]
\centering
\begin{tcolorbox}[
    tceboxstyle,
    width=\linewidth,
    left=4pt,
    right=4pt,
    top=4pt,
    bottom=4pt,
    boxsep=2pt
]
% \footnotesize
\small
\setlength{\tabcolsep}{0pt}
\renewcommand{\arraystretch}{1.08}

\begin{tabularx}{\linewidth}{@{}>{\raggedright\arraybackslash}X@{}}

$\mathbf{TCE}=\langle \texttt{instance\_id}, \Omega_{target}, \Sigma_{init}, \Sigma_{final}, G_{adj} \rangle$ \\[3pt]

$\texttt{instance\_id}=\texttt{"swan\_cfg\_01"}$ \\[3pt]

% $\Omega_{target}=\{\,V:[(0,0),(4,2\sqrt{2}),\ldots],\; 
% E:[(v_0,v_1),\ldots]\,\}$ \\[3pt]
$\Omega_{target}=\{\,V:[(0,0),(4,2\sqrt{2}),\ldots],$ \\
\hspace*{1.6cm}$E:[(v_0,v_1),\ldots]\,\}$ \\[3pt]

$\Sigma_{init}=\{\texttt{"pieces"}:[\texttt{"LT\_1"},\texttt{"LT\_2"},\ldots]\}$ \\[3pt]

$\Sigma_{final}=$\texttt{\{}
% \quad\quad \texttt{"final\_state": \{} \\
\texttt{"final\_state": \{}
\texttt{"pieces": [} 
\texttt{\{} \\
\quad\quad\quad\quad \texttt{"piece\_id": "LT\_1",} \\
\quad\quad\quad\quad \texttt{"piece\_type": "Large\_Triangle",} \\
\quad\quad\quad\quad\texttt{"vertices": [}
\texttt{\{"coords": ["$\sqrt{2}$","$0$"]\},} \\
\quad\quad\quad\quad\quad\quad\quad\quad\quad\quad\quad\quad\quad\quad\quad\quad\quad\quad\quad\texttt{\ldots} 
\texttt{],} \\
% \quad\quad\quad\quad\quad\quad\quad\quad\quad\quad\quad\quad\quad\texttt{"$0$"]\},\ldots} 
% \texttt{],} \\
\quad\quad\quad\quad \texttt{"edges": [}
\texttt{\{"coords": [["$1$", "$0$"],} \\
\quad\quad\quad\quad\quad\quad\quad\quad\quad\quad\quad\quad\quad\quad\quad\texttt{["$0$", "$1$"]],} \\
\quad\quad\quad\quad\quad\quad\quad\quad
\quad\quad\texttt{"length": "$2$"\}, \ldots} 
\texttt{]} \\
\quad\quad\quad\quad \texttt{\}, \ldots} 
\texttt{]} 
\texttt{\}} 
\texttt{\}} \\[3pt]
$G_{adj}=\{\,(\texttt{LT\_1},\texttt{SQ\_1}),
(\texttt{SQ\_1},\texttt{LT\_2}),\ldots\,\}$

\end{tabularx}
\end{tcolorbox}
\caption{Compact representation of a TCE instance.}
\label{fig:tce_definition}
\end{figure}

\paragraph{Task 1: Outline Prediction}
% 这里主要讲任务的输入输出是什么，简单提一句如何评估，具体评估方案放到后面
This setting tests whether MLLMs can synthesize a global visual shape from local component arrangements. Rather than performing conventional visual recognition, the model is required to mentally compose the union of seven discrete tangram pieces and reason about their resulting overall shape. The input consists of the textual TCE specification of the seven tangram pieces (including piece types and exact vertex coordinates) and a candidate image displaying four distinct silhouette options (A--D). The model is required to select the correct matching silhouette from the candidate set. Accuracy and Invalid Rate are used as the evaluation metrics.

% The input provides the final spatial configuration of the seven tangram pieces, comprising their piece types and vertex coordinates. Additionally, a candidate image displaying four distinct silhouette options is provided. Given the spatial arrangement, the model infers the global outline and selects the correct silhouette from four candidates (A–D). Accuracy and Invalid Rate are used as the evaluation metrics.

\begin{figure*}
    \centering
    \includegraphics[width=\linewidth]{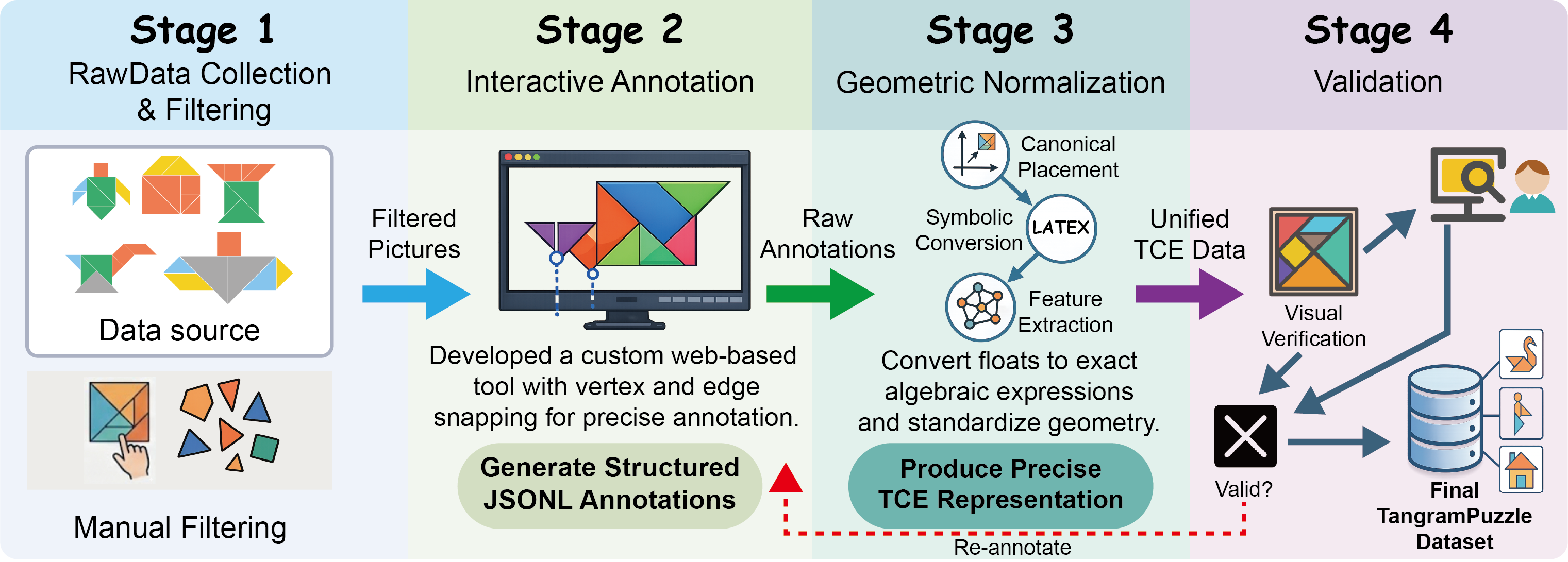}
    \caption{Overview of the TangramPuzzle data construction pipeline.}
    \label{fig:pipeline}
\end{figure*}

\paragraph{Task 2: End-to-End Tangram Solution Generation}
% This task assesses the model's ability to solve the tangram puzzle under strict geometric constraints and generate a precise executable solution. It represents a complex inverse spatial reasoning problem where the model must decompose a given target silhouette into the seven fixed rigid shapes. The input consists of a visual image of the target outline with vertex coordinates explicitly annotated within the figure. To precisely define the boundary conditions, the model is also provided with textual geometric data, including the vertex coordinates of the target outline and its edge relationships, specifying which vertices connect to form edges and their respective lengths. Furthermore, the input includes the initial state specifications for the seven tangram pieces, detailing their types, initial vertex coordinates, and constituent edges. The model is required to output the complete Tangram Configuration Expression (TCE) in JSON format. This code must explicitly define the final state of all seven pieces, including their types, vertex coordinates, and edge relationships, ensuring they perfectly fill the target outline. The evaluation checks both the geometric validity (satisfaction of rigid body and non-overlap constraints) and the shape fidelity of the generated solution, measured by Intersection over Union (IoU) and Hausdorff Distance ($d_H$) against the ground truth.

This task assesses the model's ability to solve a tangram puzzle under strict geometric constraints by generating a precise executable solution. Given a target silhouette image with annotated vertex coordinates, the model must decompose the outline into the seven fixed rigid pieces. The input also includes textual geometric data for the target outline, including vertex coordinates, edge relationships, and edge lengths, as well as the initial state of each tangram piece, including its type, vertices, and edges. The model is required to output a complete Tangram Configuration Expression (TCE) in JSON format, specifying the final state of all seven pieces so that they exactly fill the target outline. Evaluation checks both geometric validity, including rigid-body preservation and non-overlap, and shape fidelity, measured by Intersection over Union (IoU) and Hausdorff Distance ($d_H$).

\subsection{Data Construction}
To ensure high-quality geometric data, we implemented a multi-stage construction pipeline, as illustrated in Figure~\ref{fig:pipeline}, comprising source filtering, interactive annotation, and symbolic refinement.

\subsubsection{Raw Data Collection}
% We utilize the visual patterns from the KiloGram dataset as our foundational source. While KiloGram~\cite{ji2022abstract} provides a rich variety of tangram shapes, not all adhere to the strict topological constraints required for mathematical verification. Consequently, we perform a manual filtering process to exclude invalid configurations. Specifically, we remove patterns containing holes, disconnected pieces, or incomplete assemblies. Only configurations that form a single connected shape without internal voids are retained for further annotation.

% We utilize the visual patterns from the KiloGram~\cite{ji2022abstract} dataset as our foundational source. Although KiloGram provides a diverse collection of tangram compositions, TangramPuzzle differs fundamentally from it in task design. KiloGram primarily focuses on linguistic descriptions and semantic part labels (e.g., ``a person wearing a robe''), whereas TangramPuzzle introduces TCE, which grounds each shape in exact LaTeX-based algebraic coordinates and requires models to reason over precise spatial configurations rather than rely on approximate semantic categories. In addition, not all KiloGram instances satisfy the strict topological constraints required for mathematical verification. Consequently, we perform a manual filtering process to exclude invalid configurations, including patterns with holes, disconnected pieces, or incomplete assemblies. Only configurations that form a single connected shape without internal voids are retained for further comprehensive re-annotation and canonicalization.

We use visual patterns from the KiloGram~\cite{ji2022abstract} dataset as the foundational source. While KiloGram provides diverse tangram compositions, it primarily focuses on linguistic descriptions and semantic part labels (e.g., ``a person wearing a robe''). In contrast, TangramPuzzle introduces TCE to ground each shape in exact LaTeX-based algebraic coordinates, requiring models to reason over precise spatial configurations rather than approximate semantic categories. Since some KiloGram instances violate the topological constraints needed for mathematical verification, we manually filter out invalid configurations, including patterns with holes, disconnected pieces, or incomplete assemblies. Only connected shapes without internal voids are retained for re-annotation and canonicalization.

\subsubsection{Annotation Tool Development}
% Commonly used geometric software has proven insufficient for efficient and precise tangram annotation. Tools like GeoGebra\footnote{\url{https://www.geogebra.org/classic}} require users to compute coordinates before drawing and offer limited support for interactive manipulation, while manipulation-friendly platforms like Polypad\footnote{\url{https://polypad.amplify.com/p\#tangram}} lack the functionality to export precise vertex coordinates. To address these limitations, we developed a bespoke web-based tangram annotation interface. To reduce human-induced alignment errors, we further incorporate vertex snapping and edge snapping mechanisms that automatically align coincident geometric features during assembly. The tool records the coordinates of all vertices in real time and exports complete annotations directly in a structured jsonl format, substantially accelerating the labeling process. Detailed human annotation procedures and the tool are provided in Appendix~\ref{annotation}.

Existing geometric tools are insufficient for efficient and precise tangram annotation, as GeoGebra\footnote{\url{https://www.geogebra.org/classic}} requires users to compute coordinates before drawing, while manipulation-friendly platforms like Polypad\footnote{\url{https://polypad.amplify.com/p\#tangram}} cannot export precise vertex coordinates. We therefore developed a bespoke web-based annotation interface with vertex- and edge-snapping mechanisms to reduce human-induced alignment errors. The tool records all vertex coordinates in real time and exports complete annotations in structured jsonl format, substantially accelerating the labeling process. Detailed annotation procedures are provided in Appendix~\ref{annotation}.

% \subsubsection{Geometric Normalization}
% The raw annotations produced by the tool are stored in floating-point form, which is insufficient for exact geometric reasoning. To avoid ambiguity from finite-precision decimals, we post-process to normalize and convert all geometric values into exact symbolic forms. Specifically, each assembled configuration is translated such that its lower boundary lies on the $x$-axis and its left boundary lies on the $y$-axis, yielding a canonical placement. We then apply fixed rules to map all vertex coordinates to exact algebraic expressions in LaTeX. Based on these coordinates, we further extract the global outer contour, edge relations, and piece adjacency, and serialize the result into the unified TCE representation. Details on TCE construction and TCE-based generation of Task~1 multiple-choice options are provided in Appendix~\ref{Normalization}.

\subsubsection{Geometric Normalization}
The raw annotations produced by the tool are stored in floating-point form, which is insufficient for exact geometric reasoning. To avoid ambiguity from finite-precision decimals, we post-process to normalize and convert all geometric values into exact symbolic forms. Specifically, each assembled configuration is translated such that its lower boundary lies on the $x$-axis and its left boundary lies on the $y$-axis, yielding a canonical placement. For the set of raw vertices $\mathcal{V}=\{(x_i,y_i)\}_{i=1}^{N}$, this translation is defined as
\begin{equation}
\begin{aligned}
x_{\min}&=\min_{(x,y)\in\mathcal{V}}x, \qquad
y_{\min}=\min_{(x,y)\in\mathcal{V}}y,\\
(x_i',y_i')&=(x_i-x_{\min},\,y_i-y_{\min}).
\end{aligned}
\end{equation}
We then apply fixed rules to map all vertex coordinates to exact algebraic expressions in LaTeX. Based on these coordinates, we further extract the global outer contour, edge relations, and piece adjacency, and serialize the result into the unified TCE representation. Detailed TCE construction is provided in Appendix~\ref{Normalization}.

To generate samples for the task 1, we leverage the geometric properties of the TCE. For each tangram instance, we first reconstruct the ground-truth silhouette accurately based on the corresponding information encoded in the TCE. To construct the multiple-choice options, we randomly sample three distinct silhouettes from the rest of the dataset as distractors. These distractors are combined with the ground-truth silhouette, randomly shuffled, and rendered into a single candidate image labeled A through D. This automated pipeline ensures that the ground-truth silhouette is geometrically consistent with the coordinate-based input while maintaining visual diversity among the options.

\subsubsection{Human Filtering}
To ensure data correctness, we perform careful manual verification through visualization. Each annotated instance is rendered to display the assembled configuration, highlighting individual pieces and their boundaries. Instances exhibiting misaligned vertices, unintended gaps between adjacent edges, or overlaps between pieces are promptly flagged as invalid. Such cases are re-annotated using the tool, followed by repeated verification and conversion. This iterative process continues until all retained instances satisfy the geometric validity constraints required by TCE.

% Through this combination of interactive annotation, symbolic post-processing, and human-in-the-loop validation, we obtain a high-quality tangram dataset suitable for rigorous spatial reasoning evaluation.

\subsubsection{Data Statistics}
TangramPuzzle exhibits both rich structural diversity and strict geometric rigor. The target silhouettes cover a wide range of semantic categories, including abstract shapes, everyday objects, and animals, spanning a broad and challenging spectrum of difficulty from simple, compact forms to complex, articulated silhouettes with intricate boundaries. In total, the dataset contains 668 unique tangram configurations. Each configuration is used to instantiate both Task~1 and Task~2, resulting in 1,336 problem instances overall.

% 在这里详细写几种数学定理，约束条件，如何计算
\subsection{Evaluation Metrics}
\paragraph{Task 1: Outline Prediction}
To evaluate vision-grounded spatial perception, we measure the standard classification \textbf{Accuracy (Acc)}, which calculates the percentage of instances where the model selects the correct target outline. Additionally, to assess the model’s instruction-following capability and robustness, we report the \textbf{Invalid Rate}, defined as the proportion of responses that fail to map to any valid candidate option. Invalid outputs may arise from formatting errors or from generating content unrelated to the candidate options, reflecting common failure modes in instruction execution, in which case no interpretable decision can be extracted. Detailed metric computations for Task~1 are provided in Appendix~\ref{task1metrics}.

\paragraph{Task 2: End-to-End Solution Generation}

Evaluating assembly generation is non-trivial because a single target silhouette may admit multiple valid internal arrangements, making direct comparison with a single reference solution unreasonable. Moreover, relying solely on a binary success metric is insufficient to understand model behavior: a prediction may fail to produce a fully valid assembly for many different reasons. To better diagnose these failure modes, we introduce a two-stage hierarchical constraint-based evaluation framework that not only measures whether the final objective is achieved, but also reveals where and how failures occur, while providing a smoother view of progress toward valid assembly generation. Detailed metric definitions and computation procedures for Task~2 are provided in Appendix~\ref{task2metrics}. Concretely, the framework comprises:

\textbf{Stage 1: Constraint Validation.} We first check whether the generated assembly satisfies the required physical and geometric constraints by identifying three types of errors:
\begin{itemize}
    \item \textbf{Syntax Error (TSE)} checks whether the output conforms to the TCE format and piece count. Let $\mathcal{P}$ denote the predicted piece set:
    \begin{equation}
    \mathrm{TSE}=\neg\mathrm{Valid}_{\mathrm{syn}}(\mathcal{P}).
    \end{equation}

    \item \textbf{Rigid Geometry Error (RGE)} verifies strict piece-shape preservation by comparing the area, perimeter, and complete pairwise vertex-distance matrix of each predicted piece with its corresponding canonical counterpart. For a piece $P_i$ with ordered vertices $\{\mathbf{v}_1,\ldots,\mathbf{v}_n\}$ and $\mathbf{v}_{n+1}=\mathbf{v}_1$,
    \begin{equation}
    \mathrm{Area}(P_i)=
    \frac{1}{2}\left|
    \sum_{k=1}^{n}(x_k y_{k+1}-x_{k+1}y_k)
    \right|.
    \end{equation}

    \item \textbf{Physical Error (PE)} detects if any two pieces exhibit impermissible overlap or if the global union of all pieces fails to form a single connected component.
\end{itemize}
The metric \textbf{Validation Pass Rate (VPR)} reports the percentage of evaluated samples that satisfy all these constraints.

\textbf{Stage 2: Shape Similarity.} We quantify how well the constructed shape matches the target silhouette using two metrics:
\begin{itemize}
    \item \textbf{Intersection over Union (IoU)} measures the global area overlap between the predicted union $U$ and target $\mathcal{T}$:
    \begin{equation}
    \mathrm{IoU}(U,\mathcal{T})=
    \frac{\mu(U\cap\mathcal{T})}{\mu(U\cup\mathcal{T})}.
    \end{equation}

    \item \textbf{Hausdorff Distance ($d_H$)} captures fine-grained boundary deviations between shapes to penalize shape outliers:
    \begin{equation}
    \begin{aligned}
    d_H(\partial U,\partial\mathcal{T})=\max\{&
    \sup_{x\in\partial U}\inf_{y\in\partial\mathcal{T}}\|x-y\|_2,\\
    &\sup_{y\in\partial\mathcal{T}}\inf_{x\in\partial U}\|x-y\|_2\}.
    \end{aligned}
    \end{equation}
\end{itemize}
Finally, \textbf{Success} reports the percentage of samples with a fully valid assembly that solve the puzzle.

\section{Experiment}
\subsection{Experimental Setting}
\paragraph{Baselines}
% 一段话讲baseline，不用挨个写很详细
We evaluate TangramPuzzle across a diverse set of models. For open-source models, we evaluate Qwen3-VL-8B-Instruct, Qwen3-VL-32B-Instruct~\cite{bai2025qwen3vltechnicalreport}, InternVL3-78B~\cite{zhu2025internvl3}, DeepSeek-OCR~\cite{wei2025deepseek}, DeepSeek-VL2~\cite{wu2024deepseek}, and GLM-4.6V~\cite{zai2025glm46v}. For commercial models, we include GPT-5.2~\cite{openai2025gpt52}, Gemini3-Pro~\cite{deepmind2025gemini3pro}, and Claude-Sonnet-4.5~\cite{anthropic2025claudesonnet45}. Additional details are provided in Appendix~\ref{app:training_data_considerations}.

\paragraph{Details}
% 指标，实现细节等
All experiments are conducted by calling the corresponding model APIs. 
We run the evaluation pipeline on a workstation equipped with an NVIDIA RTX 2080 Ti GPU, and use a unified implementation to send requests, parse outputs, and compute all metrics across models. Prompt templates are provided in Appendix~\ref{app:prompt_templates}.

\subsection{Main Results}
% 总表，不同的模型+不同的任务与对应的指标
% 再考虑选取模型做一些别的实验，例如增加例子做ICL等，可以参考折纸的论文设计了哪些实验

% The results for Task~1 are summarized in Table~\ref{tab:task1_scores}, revealing substantial performance differences across models. This task assesses the model's ability to mentally compose a silhouette from disjointed parts and identify the correct visual match accurately. Among open-source baselines, Qwen3-VL-32B-Instruct stands out with an accuracy of 73.05\%, approaching the performance of the closed-source model GPT-5.2, while smaller or earlier architectures struggle to exceed 40\% accuracy. For commercial models, Gemini3-Pro achieves near-ceiling performance with an accuracy of $98.65\%$, underscoring its superior capacity for fine-grained visual perception and holistic mental shape composition regarding global silhouettes. For most models, the rate of invalid outputs remains low, suggesting a strong ability to follow instructions and produce interpretable decisions. However, DeepSeek-OCR and DeepSeek-VL2 exhibit noticeably higher invalid rates, suggesting that under the stress of this complex spatial reasoning task, these models often fail to form a decisive conclusion or collapse into generating irrelevant content. Detailed analysis for Task 1 is provided in Appendix~\ref{app:task1_results}.

The results for Task~1 are summarized in Table~\ref{tab:task1_scores}, revealing substantial performance differences across models. The task requires models to mentally compose a silhouette from disjointed parts and identify the correct visual match, making it difficult to solve through simple memorization or superficial cues alone. While most open-source models struggle to exceed 40\% accuracy, Gemini3-Pro achieves the highest score among all models. Most models also maintain low invalid-output rates, although a few remain less stable. Detailed analysis for Task 1 is provided in Appendix~\ref{app:task1_results}.

% Table~\ref{tab:task2_results} presents the quantitative results for Task~2. Although most models demonstrate proficiency in maintaining valid JSON syntax, as reflected by relatively low TSE scores, constraint-related failures remain widespread. In particular, Rigid Geometry Errors (RGE) and Physics Errors (PE) are highly prevalent, indicating that models struggle to comply with the fundamental geometric rules of the puzzle. One notable observation is the pronounced discrepancy between silhouette quality and the Success rate. Top-tier models like Claude-Sonnet-4.5 and GPT-5.2 achieve high IoU scores yet fail to produce a single valid solution (0\% Success). Further inspection reveals that the accompanying high RGE and PE scores indicate that this apparent visual fidelity is achieved through “cheating,” namely by impermissibly distorting rigid pieces or overlapping them to force a visual match. As a result, the vast majority of evaluated models fail to produce even a single geometrically valid solution. By contrast, Gemini3-Pro stands out as the only model demonstrating robust geometric reasoning, achieving the highest validation pass rate and success score while also maintaining superior silhouette quality.

Table~\ref{tab:task2_results} presents the results for Task~2. Their generally low \textbf{TSE} scores indicate that models can usually produce outputs in the required TCE-style format, and a lightweight recovery mechanism is applied to avoid over-penalizing minor formatting errors. This suggests that the main bottleneck is not formatting, but compositional spatial reasoning under geometric constraints. In particular, high RGE and PE across most models indicate a failure to maintain rigid geometry and physical validity, despite achieving high IoU in silhouette similarity. Further inspection reveals that this apparent visual fidelity is often achieved through ``cheating,'' namely by impermissibly distorting rigid pieces or allowing overlaps to force a visual match. Detailed analysis for Task 2 is provided in Appendix~\ref{app:task2_results}.

\subsection{Detailed Analyses}
\label{sec:analysis}

\begin{table}[t]
\centering
% \small % 如果使用了 resizebox，字体大小命令通常可以省略，因为会被自动缩放
\caption{Performance comparison on Task~1.}
\label{tab:task1_scores}
% 使用 resizebox 包裹 tabular
% \linewidth 表示缩放到当前行宽（单栏则是栏宽，通栏则是页宽）
% ! 表示保持高度比例自动调整
\resizebox{\linewidth}{!}{
    \begin{tabular}{l c c}
    \toprule
    \textbf{Model}  & \textbf{Invalid} ($\downarrow$) & \textbf{Acc} ($\uparrow$)\\
    \midrule
    % \multicolumn{3}{c}{\textit{Open-source Models}} \\
    % \midrule
    Qwen3-VL-8B~\cite{bai2025qwen3vltechnicalreport}  & \textbf{00.00} & 37.57  \\
    Qwen3-VL-32B~\cite{bai2025qwen3vltechnicalreport}  & \underline{00.15} & 73.05  \\
    InternVL3-78B~\cite{zhu2025internvl3}           & 00.75 & 29.19  \\
    DeepSeek-OCR~\cite{wei2025deepseek}            & 12.87& 22.46  \\
    DeepSeek-VL2~\cite{wu2024deepseek}             & 10.18& 25.45  \\
    GLM-4.6V~\cite{zai2025glm46v}                 & \textbf{00.00} & 30.84  \\
    \midrule
    % \multicolumn{3}{c}{\textit{Closed-source Models}} \\
    % \midrule
    GPT-5.2~\cite{openai2025gpt52}                 & \textbf{00.00} & \underline{77.54}  \\
    Gemini3-Pro~\cite{deepmind2025gemini3pro}             & \textbf{00.00} & \textbf{98.65}  \\
    Claude-Sonnet-4.5~\cite{anthropic2025claudesonnet45}       & 00.45 & 51.20  \\
    \bottomrule
    \end{tabular}
}
\end{table}

\begin{table*}[!t]
\centering
\scriptsize
\caption{Performance comparison on Task~2 with different reasoning strategies.}
\label{tab:task2_results}
\resizebox{\textwidth}{!}{
\begin{tabular}{l c c c c c c c}
\toprule
\multirow{2}{*}{\textbf{Model}} & \multicolumn{3}{c}{\textbf{Constraint error} ($\downarrow$)} & \multirow{2}{*}{\textbf{VPR} ($\uparrow$)} & \multicolumn{2}{c}{\textbf{Silhouette Quality}} & \multirow{2}{*}{\textbf{Success} ($\uparrow$)} \\
\cmidrule(lr){2-4} \cmidrule(lr){6-7}
 & \textbf{TSE} ($\downarrow$) & \textbf{RGE} ($\downarrow$) & \textbf{PE} ($\downarrow$) & & \textbf{IoU} ($\uparrow$) & \textbf{Hausdorff} ($\downarrow$) & \\
\midrule
\multicolumn{8}{c}{\textit{Vanilla MLLM}} \\
\midrule
Qwen3-VL-8B-Instruct~\cite{bai2025qwen3vltechnicalreport}   & 23.35 & \underline{40.42} & \underline{80.69} & 00.00    & 20.25 & 07.81 & 00.00      \\
Qwen3-VL-32B-Instruct~\cite{bai2025qwen3vltechnicalreport}  & 07.93  & 83.08 & 94.46 & 00.00    & 50.55 & 01.39 & 00.00      \\
InternVL3-78B~\cite{zhu2025internvl3}          & 06.59  & 91.77 & 92.96 & 00.00    & 36.15 & 03.18 & 00.00      \\
GLM-4.6V~\cite{zai2025glm46v}               & \underline{06.29}  & 91.02 & 93.56 & \underline{00.15} & 43.79 & 01.37 & \underline{00.15}   \\
GPT-5.2~\cite{openai2025gpt52}                & 07.19  & 90.12 & 92.81 & 00.00    & 58.49 & 00.84 & 00.00      \\
Gemini3-Pro~\cite{deepmind2025gemini3pro}            & 08.83  & \textbf{37.57} & \textbf{65.57} & \textbf{22.60}& \textbf{85.93} & \textbf{00.37} & \textbf{21.56} \\
Claude-Sonnet-4.5~\cite{anthropic2025claudesonnet45}      & \textbf{00.60}  & 98.20 & 97.90 & \underline{00.15} & \underline{61.61} & \underline{00.57} & 00.00      \\
\midrule
\multicolumn{8}{c}{\textit{In-Context Learning (ICL)}} \\
\midrule
Qwen3-VL-8B-Instruct~\cite{bai2025qwen3vltechnicalreport}  & 33.98 & \underline{57.04} & \underline{66.17} & \underline{00.15} & 28.48 & 03.51 & \underline{00.00} \\
Qwen3-VL-32B-Instruct~\cite{bai2025qwen3vltechnicalreport} & \textbf{06.74}  & 87.28 & 93.71 & 00.00    & 48.65 & 01.49 & \underline{00.00} \\
GPT-5.2~\cite{openai2025gpt52}               & \underline{09.43}  & 86.98 & 90.72 & 00.00    & \underline{60.30} & \underline{00.74} & \underline{00.00} \\
Gemini3-Pro~\cite{deepmind2025gemini3pro}           & 10.78 & \textbf{34.28} & \textbf{61.53} & \textbf{26.20} & \textbf{87.36} & \textbf{00.35} & \textbf{24.85} \\

\midrule
\multicolumn{8}{c}{\textit{Chain-of-Thought (CoT)}} \\
\midrule
Qwen3-VL-8B-Instruct~\cite{bai2025qwen3vltechnicalreport}  & 23.35 & \underline{40.12} & \underline{78.74} & 00.00     & 18.64 & 07.93 & 00.00 \\
Qwen3-VL-32B-Instruct~\cite{bai2025qwen3vltechnicalreport} & \underline{06.29}  & 83.83 & 95.51 & \underline{00.15}  & 51.51 & 01.38 & \underline{00.15} \\
GPT-5.2~\cite{openai2025gpt52}      & \textbf{04.04}  & 91.32 & 95.96 & 00.00     & \underline{58.80} & \underline{00.68} & 00.00 \\
Gemini3-Pro~\cite{deepmind2025gemini3pro}  & 24.40 & \textbf{33.38} & \textbf{51.35} & \textbf{20.66} & \textbf{83.12} & \textbf{00.50} & \textbf{19.76} \\

%Decomposition-based Prompting
\midrule
\multicolumn{8}{c}{\textit{Decomposition-Based Strategy}} \\
\midrule
Qwen3-VL-8B-Instruct~\cite{bai2025qwen3vltechnicalreport}  & 26.35 & \underline{41.47} & 78.14 & 00.00     & 24.22 & 07.11 & \underline{00.00} \\
Qwen3-VL-32B-Instruct~\cite{bai2025qwen3vltechnicalreport} & \textbf{06.89}  & 86.23 & 94.61 & \underline{00.15}  & 51.25 & 01.23 & \underline{00.00} \\
GPT-5.2~\cite{openai2025gpt52}      & 35.33 & 60.78 & \underline{64.22} & 00.00     & \underline{60.25} & \underline{00.88} & \underline{00.00} \\
Gemini3-Pro~\cite{deepmind2025gemini3pro}  & \underline{25.45} & \textbf{30.84} & \textbf{52.69} & \textbf{20.21} & \textbf{86.26} & \textbf{00.36} & \textbf{18.86} \\

\bottomrule
\end{tabular}
}
\end{table*}

\paragraph{Effect of In-context Learning (ICL)}
% To examine whether the bottleneck in Task~2 arises from difficulties in following the TCE schema or from deeper limitations in spatial reasoning, we apply in-context learning (ICL) by prepending three solved examples to the prompt, as shown in Table~\ref{tab:task2_results}. Contrary to the expectation that examples stabilize output, we observe a trade-off: the introduction of dense symbolic contexts tends to increase syntactic errors (TSE), suggesting a cognitive load that distracts models from strict formatting requirements. However, for successfully parsed responses, ICL generally improves silhouette quality (IoU), indicating that while examples help refine shape approximation, they cannot spontaneously induce a fundamental understanding of rigid body constraints in models that lack them. A detailed analysis of in-context learning is provided in Appendix~\ref{In-context Learning Details}.

To examine whether the bottleneck in Task~2 stems from format unfamiliarity or deeper spatial reasoning limitations, we evaluate models under an In-Context Learning (ICL) setting by prepending three solved examples, as shown in Table~\ref{tab:task2_results}. With ICL, some models exhibit higher TSE rates than in the zero-shot setting, suggesting that the longer symbolic context may introduce additional cognitive load and destabilize structured generation. Detailed analysis is provided in Appendix~\ref{In-context Learning Details}.

% Rather than uniformly helping, ICL introduces a trade-off: the additional symbolic context tends to increase syntactic errors, while successfully parsed outputs often show better silhouette quality. This suggests that examples improve shape approximation but do not induce robust understanding of rigid geometric constraints. Detailed analysis is provided in Appendix~\ref{In-context Learning Details}.

\paragraph{Effect of Chain-of-Thought (CoT)}
% As shown in the ``Chain-of-Thought (CoT)'' results in Table~\ref{tab:task2_results}, 

% As shown in Table~\ref{tab:task2_results}, CoT yields only limited and highly uneven benefits. While it can improve formatting or some intermediate metrics for certain models, it does not fundamentally resolve the core difficulty of Task~2, namely generating geometrically valid assemblies under rigid and physical constraints. In particular, gains in syntax compliance do not consistently translate into better constraint satisfaction, and in some cases improved silhouette matching is accompanied by increased geometric deformation or overlap, further reinforcing the tendency of models to rely on visual cheating.'' For smaller models, performance remains largely unchanged, suggesting that without sufficient spatial grounding, simply adding a thinking step is not enough to solve the inverse geometric reasoning required by the task. Detailed analysis is presented in Appendix~\ref{CoT}.

As shown in Table~\ref{tab:task2_results}, CoT provides only limited and uneven benefits. While it slightly improves formatting or some intermediate metrics for certain models, it does not resolve the core difficulty of Task~2. In some cases, better silhouette matching even comes with increased geometric violations, suggesting continued reliance on ``visual cheating.'' For smaller models, performance remains largely unchanged, indicating that simply adding step-by-step reasoning is insufficient without stronger spatial grounding. Detailed analysis is provided in Appendix~\ref{CoT}.

\paragraph{Effect of Systematic Decomposition}
% As shown in the ``Decomposition-Based Strategy'' results in Table~\ref{tab:task2_results}

% Motivated by the human expert analysis in Appendix~\ref{app:human_perf}, we further evaluate a decomposition-based strategy for Task~2, where models are encouraged to view both the target silhouette and the tangram pieces as compositions of the same basic unit, namely the small isosceles right triangle. As shown in Table~\ref{tab:task2_results}, this strategy leads to modest but generally consistent improvements in silhouette quality, and for some stronger models also reduces rigid-geometry and physical errors, suggesting that a shared geometric basis can help with global shape planning and constraint awareness. However, these gains do not consistently translate into higher final success, and smaller models remain largely unchanged. Overall, decomposition provides a more task-aligned reasoning prior than generic prompting, but limited compositional spatial reasoning remains the main obstacle to producing valid assemblies. Detailed analysis is presented in Appendix~\ref{Decomposition}.

Motivated by the human expert analysis in Appendix~\ref{app:human_perf}, we evaluate a decomposition-based prompting strategy for Task~2 that encourages models to represent both the target silhouette and tangram pieces using a shared unit—the small isosceles right triangle. As shown in Table~\ref{tab:task2_results}, this approach yields modest improvements in silhouette quality and slightly reduces geometric constraint violations for some models. However, these gains rarely translate into higher final Success, and smaller models show little change. Overall, while decomposition provides a more structured reasoning prior, limited compositional spatial reasoning remains the primary obstacle. Detailed analysis is provided in Appendix~\ref{Decomposition}.

\begin{table*}[!t]
\centering
\scriptsize
\caption{Performance comparison on Task~2 under different settings.}
\label{tab:task2_results2}
\resizebox{\textwidth}{!}{
\begin{tabular}{l c c c c c c c}
\toprule
\multirow{2}{*}{\textbf{Model}} & \multicolumn{3}{c}{\textbf{Constraint error} ($\downarrow$)} & \multirow{2}{*}{\textbf{VPR} ($\uparrow$)} & \multicolumn{2}{c}{\textbf{Silhouette Quality}} & \multirow{2}{*}{\textbf{Success} ($\uparrow$)} \\
\cmidrule(lr){2-4} \cmidrule(lr){6-7}
 & \textbf{TSE} ($\downarrow$) & \textbf{RGE} ($\downarrow$) & \textbf{PE} ($\downarrow$) & & \textbf{IoU} ($\uparrow$) & \textbf{Hausdorff} ($\downarrow$) & \\
\midrule
\multicolumn{8}{c}{\textit{Visual-Centric Setting}} \\
\midrule
Qwen3-VL-8B-Instruct~\cite{bai2025qwen3vltechnicalreport}  & \underline{15.42} & 17.07 & 86.23 & 00.00     & 11.39 & 12.28 & 00.00 \\
Qwen3-VL-32B-Instruct~\cite{bai2025qwen3vltechnicalreport} & \textbf{09.58}  & 74.70 & 93.26 & \underline{00.15}  & 43.69 & 02.78  & \underline{00.15} \\
DeepSeek-OCR~\cite{wei2025deepseek}          & 100.0   & \textbf{00.00}     & \textbf{01.35}  & 00.00     & 11.72 & 07.81  & 00.00 \\
DeepSeek-VL2~\cite{wu2024deepseek}          & 99.85 & \underline{00.15}  & \underline{13.32} & 00.00     & 12.08 & 05.49  & 00.00 \\
GPT-5.2~\cite{openai2025gpt52}               & 39.67 & 58.23 & 58.08 & 00.00     & \underline{53.67} & \underline{01.21}  & 00.00 \\
Gemini3-Pro~\cite{deepmind2025gemini3pro}           & 16.77 & 32.34 & 56.89 & \textbf{25.00} & \textbf{82.75} & \textbf{00.42}  & \textbf{22.75} \\

%Text-Only Setting
\midrule
\multicolumn{8}{c}{\textit{Text-Only Setting}} \\
\midrule
Qwen3-VL-8B-Instruct~\cite{bai2025qwen3vltechnicalreport}  & 50.15 & \textbf{26.35} & \underline{51.95} & \underline{00.00}     & 21.56 & 07.45 & \underline{00.00} \\
Qwen3-VL-32B-Instruct~\cite{bai2025qwen3vltechnicalreport} & \textbf{13.92} & 72.16 & 87.72 & \underline{00.00}     & 47.66 & \underline{02.20} & \underline{00.00} \\
GPT-5.2~\cite{openai2025gpt52}      & 54.04 & 40.42 & \textbf{43.86} & \underline{00.00}     & \underline{54.40} & 02.23 & \underline{00.00} \\
Gemini3-Pro~\cite{deepmind2025gemini3pro}  & \underline{28.14} & \underline{33.68} & 56.74 & \textbf{13.32} & \textbf{81.15} & \textbf{00.59} & \textbf{12.28} \\

%TCE-Light Prompting
\midrule
\multicolumn{8}{c}{\textit{TCE-Light Setting}} \\
\midrule
Qwen3-VL-8B-Instruct~\cite{bai2025qwen3vltechnicalreport}  & 19.91 & \textbf{00.00} & \underline{81.59} & \underline{00.00}     & 05.10 & 18.18 & \underline{00.00} \\
Qwen3-VL-32B-Instruct~\cite{bai2025qwen3vltechnicalreport} & \underline{04.49} & \textbf{00.00} & 95.81 & \underline{00.00}     & \underline{13.91} & \underline{16.49} & \underline{00.00} \\
GPT-5.2~\cite{openai2025gpt52}      & \textbf{01.95} & \textbf{00.00} & 98.80 & \underline{00.00}     & 07.45 & 18.35 & \underline{00.00} \\
Gemini3-Pro~\cite{deepmind2025gemini3pro}  & 14.82 & \textbf{00.00} & \textbf{80.84} & \textbf{04.49} & \textbf{36.99} & \textbf{12.18} & \textbf{03.59} \\

\bottomrule
\end{tabular}
}
\end{table*}

\paragraph{Effect of Textual Geometry}
We further investigated the models' dependency on textual metadata through an ablation study in a Visual-Centric setting, as shown in Table~\ref{tab:task2_results}. In this setup, we removed the textual description of the target outline (i.e., vertex and edge lists), providing only the outline image with annotated coordinates alongside the initial piece descriptions. Most MLLMs show a clear performance drop, accompanied by more hallucinations and syntax errors, indicating a strong dependence on textual ``crutches'' rather than precise geometric extraction from vision alone. Textual geometry analysis is detailed in Appendix~\ref{Textual}.

% The results show a significant performance drop for most MLLMs, with increased hallucinations and syntax errors. This confirms that current models largely rely on textual ``crutches'' rather than precisely extracting geometric coordinates from visual inputs. A notable exception is Gemini3-Pro, which maintained high performance, demonstrating superior capability in grounding geometric data directly from vision.

\paragraph{Effect of Removing Visual Input}
% As shown in the ``Text-Only Setting'' results in Table~\ref{tab:task2_results}

% To examine whether Task~2 degenerates into a purely symbolic encoding--decoding problem once converted into TCE, we conduct a text-only ablation in which models receive only the textual description of the target outline vertices, without the target image. As shown in Table~\ref{tab:task2_results}, removing visual input leads to a clear performance drop, most notably for Gemini3-Pro, while the other models still fail to produce valid final assemblies. This suggests that the visual silhouette provides a crucial spatial anchor that cannot be fully replaced by symbolic coordinates alone. More broadly, if the task were mainly symbolic mapping, performance would be expected to remain stable or even improve in the text-only setting; instead, the observed degradation indicates that solving TangramPuzzle requires genuine visual-spatial grounding and compositional reasoning over a large transformation space, rather than simple end-to-end symbolic transduction. Detailed analysis is presented in Appendix~\ref{Removing_Visual}.

To examine whether Task~2 degenerates into a purely symbolic encoding--decoding problem once converted into TCE, we conduct a text-only ablation in which models receive only textual descriptions of the target outline vertices. As shown in Table~\ref{tab:task2_results}, removing visual input leads to a clear performance drop, most notably for Gemini3-Pro. This suggests that the visual silhouette provides a crucial spatial anchor that cannot be fully replaced by symbolic coordinates alone. Overall, the degradation indicates that solving TangramPuzzle requires integrating both visual and symbolic modalities for spatial reasoning, rather than relying on simple symbolic mapping. Detailed analysis is presented in Appendix~\ref{Removing_Visual}.

\paragraph{Effect of TCE-Light Representation}
To examine whether Task~2 failures stem from the full TCE representation rather than spatial reasoning itself, we introduce a simplified \textit{TCE-Light} setting, where models output only one rigid transformation per piece, including \texttt{piece\_id}, rotation angle, optional reflection, and translation. The final vertices are then deterministically computed from the initial pieces, reducing symbolic coordinate generation and making RGE zero for successfully parsed outputs. As shown in Table~\ref{tab:task2_results}, however, this simplification does not solve the task: models still exhibit high PE, low VPR, and near-zero Success, with silhouette quality often deteriorating. Detailed analysis is provided in Appendix~\ref{app: TCE-Light}.

\paragraph{Relationship between Task 1 and Task 2}
To better understand the relationship between the two tasks, we compute cross-model correlations between Task~1 Accuracy and the main Task~2 metrics using both Pearson and Spearman correlations. The results show that Task~1 Accuracy correlates strongly with Task~2 IoU, suggesting that global-shape composition helps approximate the target silhouette, but it does not reliably predict fully valid assemblies. Detailed analysis and the correlation table are provided in Appendix~\ref{app:task_relation}.

\paragraph{Human Performance}
\begin{table}[t]
\centering
\caption{Human Performance on Tasks 1 and 2.}
\label{tab:human_performance}
\resizebox{\linewidth}{!}{
\begin{tabular}{l c c c}
\toprule
\textbf{Human} & \textbf{Success} ($\uparrow$) & \textbf{Complexity} ($\downarrow$) & \textbf{Time usage (min)} ($\downarrow$) \\
\midrule

\multicolumn{4}{c}{\textit{Human performance on Task~1}} \\
\midrule

Human 1 & \textbf{100.0}    & 02.10 & 03.10   \\
Human 2 & \textbf{100.0}   & \textbf{01.50} & \underline{02.80} \\
Human 3 & \textbf{100.0} & \underline{01.70} & \textbf{02.50} \\
Average        & 100.0 & 01.77   & 02.80   \\

\midrule
\multicolumn{4}{c}{\textit{Human performance on Task~2}} \\
\midrule

Human 1 & \underline{72.00}    & \underline{04.10} & \textbf{06.30}   \\
Human 2 & \textbf{100.0}   & \textbf{03.80} & \underline{06.60} \\
Human 3 & 46.00 & 04.30 & 07.10 \\
Average        & 72.67 & 04.07   & 06.67   \\

\bottomrule
\end{tabular}
}
\end{table}

We evaluate human performance on both tasks using domain experts, as shown in Table~\ref{tab:human_performance}. All participants achieve 100\% accuracy on Task~1, while Task~2 remains more challenging but rarely leads to geometric-constraint violations; human failures are typically left unsolved rather than invalid. Detailed results and solving strategies (e.g., systematic triangle decomposition) are provided in Appendix~\ref{app:human_perf}. We also conduct a complementary human evaluation of MLLM outputs on Task~2 to assess perceptual plausibility beyond strict automatic metrics, detailed in Appendix~\ref{app:Human_Evaluation}.

% (ranging from systematic triangle decomposition to iterative trial-and-error) 

% Detailed analysis is provided in Appendix~\ref{app:Human_Evaluation}.

% Detailed results and solving strategies are provided in Appendix~\ref{app:human_perf}.

\paragraph{Human Evaluation on MLLMs}
Automatic metrics primarily capture strict geometric validity and may overlook perceptual plausibility. To complement them, we evaluate Task~2 outputs using three 5-point criteria: Perceptual Silhouette Match (PSM), which measures visual similarity to the target; Piece Plausibility \& Canonicality (PPC), which assesses whether piece placements appear natural rather than strained; and Near-Miss Closeness (NMC), which indicates how close an incorrect prediction is to a valid solution. Table~\ref{tab:Human evaluation of MLLM} reports the mean $\pm$ sample standard deviation across three experts. The low inter-rater variability, with standard deviations ranging from 0.15 to 0.36, further supports the consistency of the evaluation. Gemini3-Pro achieves consistently high scores, with even unsuccessful outputs often judged close to correct. By contrast, GPT-5.2 scores substantially lower, showing more perceptual mismatches, implausible constructions, and predictions far from valid solutions.

% Automatic metrics emphasize geometric validity but may overlook perceptual plausibility. We therefore evaluate Task~2 outputs using three 5-point criteria: Perceptual Silhouette Match (PSM), measuring similarity to the target; Piece Plausibility \& Canonicality (PPC), assessing whether placements appear natural; and Near-Miss Closeness (NMC), indicating how close an incorrect prediction is to a valid solution. Table~\ref{tab:Human evaluation} reports the mean $\pm$ sample standard deviation across three experts, with low variability (0.15--0.36) supporting evaluation consistency. Gemini3-Pro scores high, with even unsuccessful outputs often judged close to correct, whereas GPT-5.2 scores lower, showing perceptual mismatches, implausible constructions, and predictions far from valid solutions.

\begin{table}[t]
\centering
\caption{Human evaluation of MLLM on Task~2.}
\label{tab:Human evaluation of MLLM}
\resizebox{\linewidth}{!}{
\begin{tabular}{l l c c c}
\toprule
\textbf{Model} & \textbf{Expert} & \textbf{PSM} ($\uparrow$) & \textbf{PPC} ($\uparrow$) & \textbf{NMC} ($\uparrow$) \\
\midrule
\multirow{4}{*}{Gemini3-Pro}
 & Expert 1 & 04.20   & 04.30 & 04.60 \\
 & Expert 2 & 04.50   & 04.60   & 04.10   \\
 & Expert 3 & 04.20 & 04.50 & 04.00 \\
 & Average   & $\bm{04.30 \pm 0.17}$   & $\bm{04.47 \pm 0.15}$   & $\bm{04.23 \pm 0.32}$   \\
\midrule
\multirow{4}{*}{GPT-5.2}
 & Expert 1 & 01.60   & 01.50 & 01.80 \\
 & Expert 2 & 01.20   & 01.10   & 01.10   \\
 & Expert 3 & 01.50 & 01.20 & 01.60 \\
 & Average   & $\underline{01.43 \pm 0.21}$   & $\underline{01.27 \pm 0.21}$   & $\underline{01.50 \pm 0.36}$   \\
\bottomrule
\end{tabular}
}
\end{table}

\subsection{Case Study}

% Figure~\ref{fig:results} visualizes the generation results for a U-shaped target (Task 2). Gemini3-Pro achieves a perfect assembly, demonstrating its ability to effectively ground coordinate constraints to solve complex compositional problems. A critical failure mode observed across other models is the tendency to prioritize visual silhouette matching at the expense of geometric constraints. For instance, Claude-Sonnet-4.5 attempts to force an alignment with the target contour by non-rigidly elongating the edges of the left and right triangles, whereas InternVL3-78B generates severe overlaps to maximize area coverage. Meanwhile, GPT-5.2 produces a visually plausible silhouette but fails to preserve component fidelity, utilizing an incorrect piece inventory (e.g., hallucinating an extra square while omitting the parallelogram). We further observe that smaller models such as Qwen3-VL-8B-Instruct often arrange pieces in a simple linear sequence, indicating an inability to map symbolic coordinate specifications onto a 2D planar workspace.

\begin{figure}
    \centering
    \includegraphics[width=\linewidth]{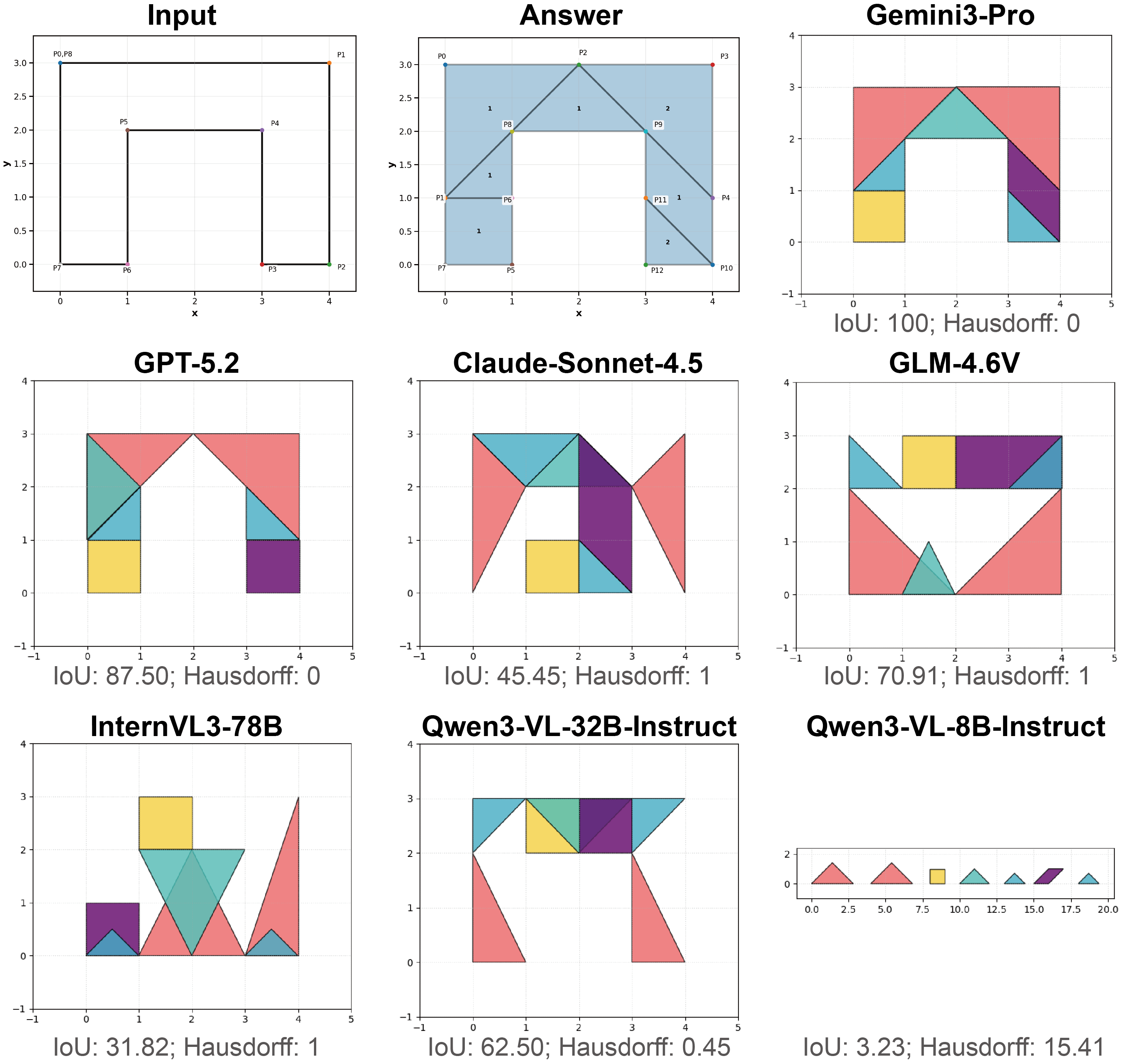}
    \caption{Visualization of a case from Task~2.}
    \label{fig:results}
\end{figure}

Figure~\ref{fig:results} presents a case study on a U-shaped target in Task~2. Gemini3-Pro is the only model that produces a fully valid assembly, showing strong geometric grounding and compositional reasoning. Other models tend to favor silhouette matching over geometric validity: Claude-Sonnet-4.5 deforms pieces, InternVL3-78B produces severe overlaps, and GPT-5.2 generates a plausible outline but with an incorrect piece inventory. Detailed analysis is presented in Appendix~\ref{app:Case_Study}.

% By contrast, other models typically favor visual silhouette matching over geometric validity: Claude-Sonnet-4.5 deforms rigid pieces, InternVL3-78B produces severe overlaps, and GPT-5.2 generates a visually plausible outline but with an incorrect piece inventory. Detailed analysis is presented in Appendix~\ref{app:Case_Study}.

% Smaller models such as Qwen3-VL-8B-Instruct often arrange pieces linearly, indicating difficulty in constructing a valid 2D assembly from symbolic geometry. 

\section{Conclusion}
We presented TangramPuzzle, a geometry-grounded benchmark specifically designed for evaluating MLLMs on compositional spatial reasoning. Addressing the limitations of existing semantic-focused evaluations, we introduced the Tangram Construction Expression (TCE), a symbolic geometric framework that enables the precise, machine-verifiable assessment of rigid body constraints and physical feasibility. The benchmark encompasses the Outline Prediction and End-to-End TCE Code Generation tasks, which jointly probe the spectrum of spatial intelligence. We conducted a comprehensive evaluation of open-source and closed-source MLLMs covering different scales, accompanied by in-depth exploratory analyses on in-context learning and modality dependency. By shifting the evaluation paradigm from approximate visual matching to exact constraint satisfaction, TangramPuzzle provides a standardized and challenging testbed for future research.

\section*{Acknowledgments}
This research is supported by the National Natural Science Foundation of China (Grant No. 62276154), the Natural Science Foundation of Guangdong Province (Grant No. 2024TQ08X729), the Basic Research Fund of Shenzhen City (Grant Nos. JCYJ20240813112009013 and GJHZ20240218113603006), and the Major Key Project of PCL for Experiments and Applications (Grant No. PCL2024A08).

\section*{Limitations}
% While TangramPuzzle establishes a rigorous framework for evaluating compositional spatial reasoning through exact geometric constraints, several limitations remain. The benchmark is currently confined to 2D planar geometry with a fixed number of components, which may not fully capture the scalability and complexity of real-world 3D spatial manipulation tasks involving occlusion or variable object counts. Future work will seek to extend this evaluation paradigm to three-dimensional settings, thereby bringing the evaluation closer to real-world embodied scenarios. Further discussion future extensions is provided in Appendix~\ref{sec:discussion}.

While TangramPuzzle provides a rigorous framework for evaluating compositional spatial reasoning under exact geometric constraints, it is currently limited to 2D planar geometry with a fixed set of components, and thus does not fully capture real-world 3D complexities such as occlusion and viewpoint changes. Future work will extend this evaluation paradigm toward richer and more realistic spatial assembly settings; further discussion and future extensions are provided in Appendix~\ref{sec:discussion}.

\section*{Ethical Considerations}
\paragraph{Potential Risks}
The TangramPuzzle benchmark focuses on abstract geometric reasoning and is constructed entirely from publicly available sources. Due to the inherently abstract nature of these geometric shapes, the dataset is devoid of personally identifiable information (PII), sensitive biometric data, or offensive content. Consequently, there are no foreseeable risks related to safety, discrimination, or surveillance.

\paragraph{Ethical Statement}
Our data construction process strictly adhered to ethical research guidelines. All source images were derived from public domains, ensuring no copyright infringement. During the human-in-the-loop annotation and validation stages, all participants were informed of the nature of the task and compensated fairly in accordance with local labor standards. Furthermore, as the dataset comprises solely rigid geometric shapes, it is inherently free from biometric data, social biases, or other privacy concerns. Thus, the proposed research direction and tasks are ethically benign and socially harmless. All data and annotations will be released under a permissive open-source license upon acceptance to facilitate transparency and reproducibility.

\paragraph{LLMs Usage Statement}
We use large language models and multimodal models as baselines to generate predictions for evaluation and analysis. During the writing of this paper, we also use large language models to assist with grammar checking and improving text fluency. The authors have carefully reviewed and verified all suggested revisions provided by the large language models to ensure their correctness, and take full responsibility for the content of this paper.

\bibliography{custom}

\appendix

\section{Comparison with Existing Spatial Benchmarks}
\label{app:benchmark_comparison}

To better contextualize our contributions, we provide a detailed comparison between TangramPuzzle and representative spatial reasoning benchmarks in Table~\ref{tab:benchmark_comparison}. Compared to prior datasets, TangramPuzzle differs in four main aspects.

\paragraph{1. From discriminative VQA to inverse generation with verifiable structure:} While recent benchmarks have significantly advanced spatial evaluation, they each emphasize different facets: OmniSpatial~\cite{jia2025omnispatial} focuses on broad coverage (spanning 4 major categories and 50 subcategories), MMSI-Bench~\cite{yang2025mmsi} highlights camera/object/region relations in multi-view real-world scenes, and LEGO-Puzzles~\cite{tang2025lego} targets 3D assembly sequences. Furthermore, prior Tangram datasets~\cite{zhang2024tangram} similarly focus on identifying and counting basic geometric elements (e.g., points, lines, circles, triangles). However, these benchmarks are predominantly formulated as Visual Question Answering (VQA) tasks. While valuable, VQA tasks can encourage shortcut behaviors (e.g., exploiting language priors) and do not require the model to construct a globally consistent structure. In contrast, our TCE framework provides a standardized, machine-verifiable symbolic bridge for compositional reasoning. By utilizing exact LaTeX algebraic expressions, we eliminate linguistic ambiguity and enable objective, automated verification.

\paragraph{2. Complementary focus to ORIGAMISPACE:} Although ORIGAMISPACE~\cite{xuorigamispace} is also explicitly generative, it fundamentally targets folding and deformation reasoning. It focuses on procedural manipulation by transforming a 2D crease pattern into a 3D folded shape through multi-step folding, and it relies heavily on an origami compiler. Additionally, these tasks often rely on task-specific answer formats or final-choice accuracy rather than exact symbolic representations of all component geometries. Conversely, TangramPuzzle requires a "global-to-local" decomposition where models must mentally simulate and inversely assemble a complete structure from scratch under strict geometric constraints. Thus, the two benchmarks evaluate complementary dimensions of spatial reasoning.

\paragraph{3. Fine-grained diagnosis to prevent ``visual cheating'':} TangramPuzzle introduces TCE as a formal representation alongside a hierarchical verifier, which precisely categorizes failures into syntax errors, rigid-geometry errors, physical errors, and silhouette mismatches. This provides much deeper diagnostic value than a single accuracy score. For instance, our experiments reveal that models sometimes assemble visually plausible silhouettes by violating piece rigidity or allowing overlaps. This type of "visual cheating" is easily overlooked in benchmarks that only evaluate semantic answers or coarse visual similarities.

\paragraph{4. Controlled 2D environment to isolate core spatial reasoning:} While 3D tasks add realism, they inevitably introduce confounds such as viewpoint variation, occlusion, depth ambiguity, and simulator assumptions. TangramPuzzle intentionally adopts a minimal yet expressive 2D abstraction governed by strict Euclidean and topological constraints. This design choice directly isolates the core challenge of compositional spatial reasoning, making model failures clear, reproducible, and automatically verifiable.

\section{Detailed Definition of Tangram Construction Expression (TCE)}
\label{app:tce_definition}

The Tangram Construction Expression (TCE) represents each tangram instance in a structured and machine-verifiable form. It consists of five core components: instance\_id, target\_outline, initial\_state, final\_state, and adjacency\_graph. Their detailed definitions are given below.

The field \textbf{instance\_id} uniquely identifies each puzzle instance and serves as a stable key for indexing, retrieval, and evaluation. Both \textbf{initial\_state} and \textbf{final\_state} describe the geometric states of the seven tangram pieces before and after assembly. In TCE, each piece is specified by its piece type, a set of vertex coordinates, explicit edge relations, and a center point. In addition, final\_state includes a transform\_matrix that parameterizes the rigid motion applied to each piece to reach the assembled configuration, including translation, rotation angle, and optional reflection. Because tangram geometries frequently involve irrational numbers, finite decimal representations induce precision loss. To ensure geometric rigor, TCE encodes all quantities as exact algebraic expressions in LaTeX. The \textbf{target\_outline} specifies the global silhouette produced by the assembled tangram. It is represented by the outline's vertex coordinates together with its edge relations. The \textbf{adjacency\_graph} captures local interactions between pieces in the final configuration, indicating which pieces are adjacent.

\section{Human Annotation Details}
\label{annotation}
\paragraph{Recruitment and Payment.}
To ensure the high quality and rigorous geometric precision of the TangramPuzzle benchmark, we invited researchers with backgrounds in geometry and computer science to conduct the data annotation and verification. We strictly adhered to local labor standards regarding compensation. The payment rate was determined based on the estimated time required to complete the annotation for each puzzle, ensuring that the hourly rate met or exceeded the local minimum wage requirements.

\begin{figure*}
    \centering
    \includegraphics[width=\linewidth]{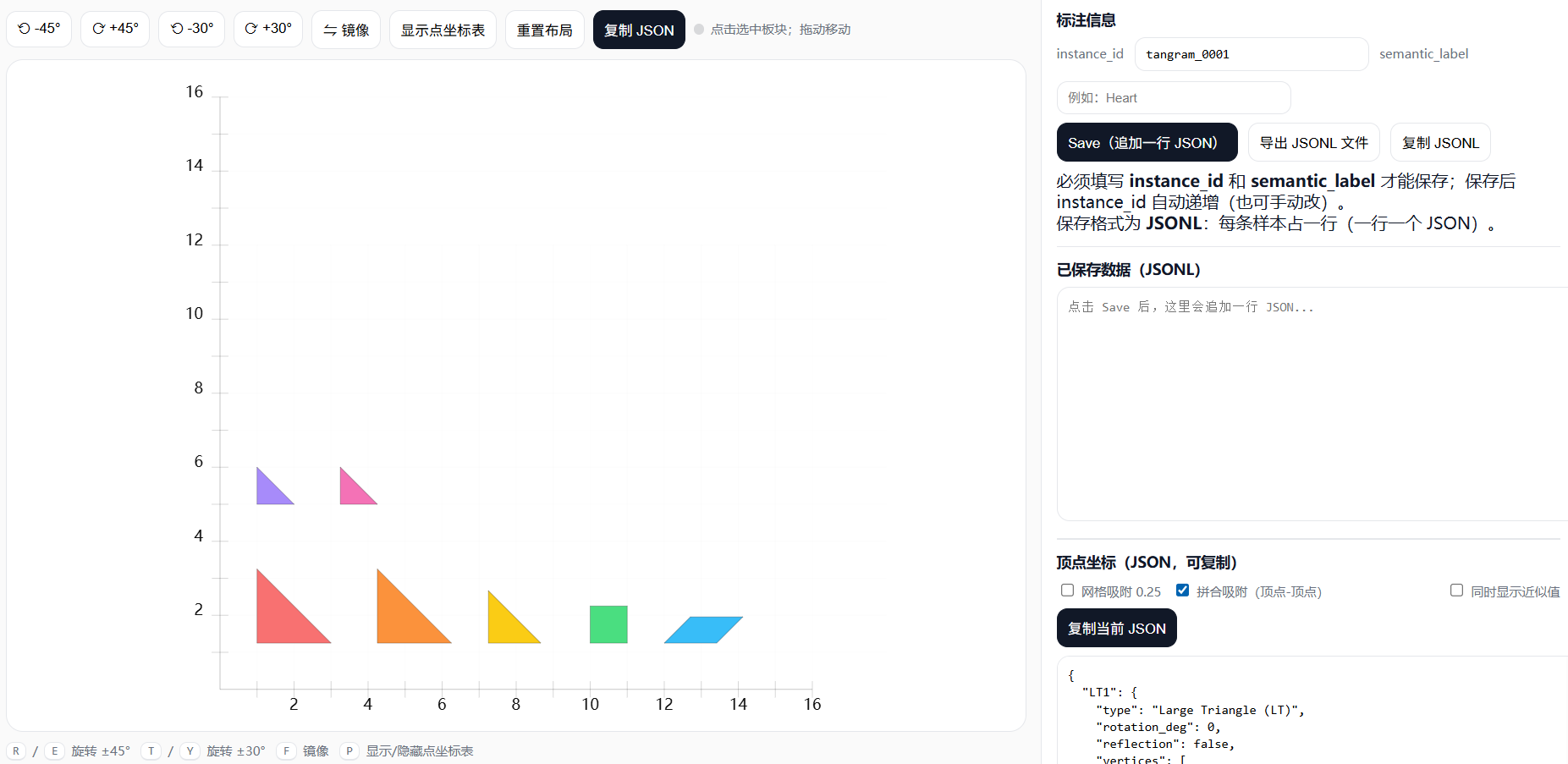}
    \caption{Annotation interface for TangramPuzzle.}
    \label{fig:annotation interface}
\end{figure*}

\paragraph{Annotation Process.}
Annotators used our interactive tangram tool to assemble each target configuration by dragging, rotating, and flipping pieces. The annotation interface is shown in Figure~\ref{fig:annotation interface}. To avoid pixel-level misalignment, the tool incorporates a magnetic snapping mechanism that automatically aligns piece vertices to the global coordinate grid or to vertices of adjacent pieces, ensuring that spatial relationships are mathematically exact rather than visually approximate. After completing an instance, annotators exported the generated data, and we conducted additional visual inspection and re-annotation when necessary (e.g., minor misalignment, unintended overlaps, or disconnected components) to ensure geometric correctness.

\paragraph{Instructions and Consent.}
Prior to the data collection process, all annotators were provided with comprehensive instructions detailing the usage of the annotation interface, the definition of the Tangram Construction Expression (TCE), and the strict physical constraints (e.g., non-overlap, boundary adherence). We obtained informed consent from all participants, who were explicitly informed that the annotated data would be used for academic research and released to the public.
% Prior to the data collection process, all annotators were provided with comprehensive instructions detailing the usage of the annotation interface, the definition of the Tangram Construction Expression (TCE), and the strict physical constraints (e.g., non-overlap, boundary adherence). To ensure Data Consent, we obtained informed agreement from all participants, having explicitly informed them via these instructions that the annotated data would be utilized for academic research and subsequently released to the public.

\section{Detailed Geometric Normalization}
\label{Normalization}

To ensure the resulting dataset is machine-verifiable and free from floating-point ambiguity, we implemented a rigorous post-processing pipeline to convert raw manual annotations into the symbolic TCE format. This process consists of three key steps:

\paragraph{1. Reference Alignment and Raw Capture.}
First, annotators utilize our custom web-based tool to manually reconstruct the tangram patterns from the source KiloGram dataset~\cite{ji2022abstract}. By dragging, rotating, and snapping the seven virtual pieces, annotators align the pieces to visually match the reference image. Upon saving, the tool records the raw floating-point coordinates of all vertices for the seven constituent pieces in their current assembled state.

\paragraph{2. Canonical Placement.}
To standardize the coordinate system across all instances and eliminate arbitrary canvas positioning, we apply a translation operation to the raw assembly. Let $\mathcal{V} = \{(x_i, y_i)\}_{i=1}^{N}$ denote the set of all vertex coordinates in the raw annotation. We compute the global minima of the assembly:
\begin{equation}
    x_{min} = \min_{(x, y) \in \mathcal{V}} x, \quad y_{min} = \min_{(x, y) \in \mathcal{V}} y
\end{equation}
We then translate every vertex $(x, y)$ to a new coordinate $(x', y')$:
\begin{equation}
    x' = x - x_{min}, \quad y' = y - y_{min}
\end{equation}
This operation ensures that the assembly's lower boundary lies strictly on the $x$-axis and its left boundary lies on the $y$-axis, resulting in a standardized, positive-quadrant layout.

\paragraph{3. Symbolic Conversion and TCE Generation.}
To support exact geometric reasoning and eliminate floating-point ambiguity, we convert the normalized coordinates into precise algebraic expressions represented in LaTeX format.

We employ a deterministic mapping algorithm to convert decimal approximations into exact rational or irrational numbers. For instance, a coordinate $0.5$ is converted to \texttt{\textbackslash frac\{1\}\{2\}}, and irrational numbers resulting from $45^\circ$ rotations (e.g., $1.414\dots$) are resolved to their symbolic forms (e.g., \texttt{\textbackslash sqrt\{2\}}). This ensures exact geometric relations (e.g., edge equality).

With the exact coordinates of the seven pieces established (forming the \texttt{final\_state}), we computationally extract the global geometry. This includes calculating the geometric union of the pieces to extract the \texttt{target\_outline} (effectively dissolving internal edges) and generating an adjacency graph by identifying shared vertex pairs and overlapping edge segments. Finally, these components are serialized into the unified TCE JSON format, providing a mathematically perfect ground truth for model evaluation.

\paragraph{4. Task 1 Multiple-Choice Construction.}
To generate samples for Task~1, we leverage the geometric information encoded in TCE. For each tangram instance, we reconstruct the ground-truth silhouette from its TCE annotation. We then construct the multiple-choice options by randomly sampling three distinct silhouettes from the remaining dataset as distractors. These distractors are combined with the ground-truth silhouette, randomly shuffled, and rendered into a single candidate image labeled A through D. This automated pipeline ensures that the correct silhouette is geometrically consistent with the coordinate-based input while maintaining visual diversity among the options.

% This automated pipeline ensures that the correct silhouette is geometrically consistent with the coordinate-based input while maintaining visual diversity among the options.

\section{Detailed Evaluation Protocols}
\label{metrics}
This section provides the formal definitions and precise calculation details for the metrics used in our evaluation.

\subsection{Task 1 Metrics}
\label{task1metrics}
\paragraph{Accuracy (Acc)}
Performance is measured using classification accuracy:
\begin{equation}
    \mathrm{Acc} = \frac{1}{N} \sum_{i=1}^{N} \mathbb{I}\!\left(\hat{a}_i = a_i\right),
\end{equation}
where $a_i \in \mathcal{A}$ denotes the ground-truth option for the $i$-th instance, $\hat{a}_i \in \mathcal{A}$ is the model's predicted option, $\mathcal{A}$ is the option set (e.g., $\{A,B,C,D\}$), and $\mathbb{I}(\cdot)$ is the indicator function.

\paragraph{Invalid Rate}
We further define the Invalid rate as the proportion of instances in which the model output does not yield any valid option that can be mapped to the candidate set. Invalid outputs may arise from formatting errors or from generating content unrelated to the candidate options. Formally:
\begin{equation}
\mathrm{Invalid} = \frac{1}{N} \sum_{i=1}^{N} \mathbb{I}\!\left(\hat{a}_i \notin \mathcal{A}\right).
\end{equation}

\subsection{Task 2 Metrics}
\label{task2metrics}
Let $\mathcal{P} = \{P_1, \dots, P_7\}$ denote the set of predicted piece polygons after assembly, and let $\mathcal{T}$ denote the ground-truth target silhouette. The evaluation pipeline consists of two stages: constraint validation and silhouette quality assessment.

\paragraph{Constraint Validation}
We consider three types of constraint errors:

\begin{itemize}
\item \textbf{TSE (Syntax Error)} checks whether the predicted output conforms to the required TCE format. A prediction is flagged as TSE if the JSON cannot be parsed, the number of pieces is incorrect, or piece types are invalid:
\begin{equation}
\mathrm{TSE} = \neg \mathrm{Valid}_{\mathrm{syn}}(\mathcal{P}),
\end{equation}
where $\mathrm{Valid}_{\mathrm{syn}}(\mathcal{P})$ denotes syntactic validity under the TCE schema.

\item \textbf{RGE (Rigid Geometry Error)} verifies that each predicted piece preserves its original shape. For each piece, we compare its area and perimeter against the corresponding ground-truth specification. Let $P_i$ be a simple polygon with ordered vertices $\{ \mathbf{v}_1, \dots, \mathbf{v}_n \}$.
Its area is computed as:
\begin{equation}
    \mathrm{Area}(P_i) = \frac{1}{2} \left| \sum_{k=1}^{n} (x_k y_{k+1} - x_{k+1} y_k) \right|,
\end{equation}
and its perimeter is given by
\begin{equation}
    \mathrm{Perimeter}(P_i) = \sum_{k=1}^{n} \left\| \mathbf{v}_{k+1} - \mathbf{v}_k \right\|_2,
\end{equation}
where $\mathbf{v}_{n+1} = \mathbf{v}_1$. A rigid geometry error is triggered if either quantity deviates from that of the corresponding canonical tangram piece.

\item \textbf{PE (Physical Error)} evaluates physical feasibility. A prediction is considered invalid if any two pieces overlap or if the union of all pieces is not a single connected shape.
\end{itemize}

\paragraph{Shape Similarity}
We adopt two complementary shape similarity metrics:

\begin{itemize}
\item \textbf{IoU (Intersection over Union)} measures the overlap between the predicted assembly $U$ and the target silhouette $\mathcal{T}$:
\begin{equation}
\mathrm{IoU}(U, \mathcal{T}) =
\frac{\mu(U \cap \mathcal{T})}{\mu(U \cup \mathcal{T})},
\end{equation}
where $\mu(\cdot)$ denotes planar area.

\item \textbf{Hausdorff Distance ($d_H$)}
To capture boundary-level deviations, we compute the Hausdorff distance between the boundaries of $U$ and $\mathcal{T}$:
\begin{multline}
d_H(\partial U, \partial \mathcal{T}) = \\
\max \left\{
\begin{aligned}
&\sup_{x \in \partial U} \inf_{y \in \partial \mathcal{T}} \|x - y\|_2, \\
&\sup_{y \in \partial \mathcal{T}} \inf_{x \in \partial U} \|x - y\|_2
\end{aligned}
\right\}.
\end{multline}
Here $\partial U$ and $\partial \mathcal{T}$ denote the boundaries of the predicted assembly and the target silhouette, respectively, and $\|\cdot\|_2$ represents the standard Euclidean distance.
\end{itemize}

\section{Details on Baselines}
\label{app:training_data_considerations}
We utilize the official chat templates and default generation parameters for all models to ensure a fair comparison.

To contextualize the experimental results, we examine the released model cards and technical reports of the open-source baselines evaluated in this work. These models are generally trained on multimodal mixtures, including general image understanding, OCR, document question answering, chart interpretation, and conventional spatial-relation VQA. However, to the best of our knowledge, their released training descriptions report no training data in the same format as our TCE-style tangram inverse-assembly task, which requires generating exact algebraic coordinates under rigid-body, non-overlap, and connectivity constraints.

For closed-source models, such as Gemini3-Pro, GPT-5.2, and Claude-Sonnet-4.5, the exact training corpora are not publicly available, and we therefore cannot rule out possible exposure to tangram-like visual patterns. Nevertheless, our Tangram Construction Expression (TCE) is newly designed for this benchmark and represents tangram assemblies as strict, machine-verifiable JSON structures with exact symbolic coordinates. This makes direct memorization of TCE-formatted solutions unlikely, even if some models may have encountered standard tangram images or natural-language tangram descriptions during pre-training.

Therefore, the results should be interpreted not as a claim about training-data purity for every model, but as evidence that current MLLMs, especially those trained mainly on semantic recognition and coarse spatial relations, still struggle with constraint-based compositional spatial reasoning.

\section{Prompt Templates}
\label{app:prompt_templates}

To ensure a fair comparison, all prompts used in our experiments are setting-specific but model-agnostic.
Within each experimental setting, all models are evaluated using exactly the same prompt template, with no model-specific prompt engineering or optimization.
The placeholders in the following listings, such as \texttt{\detokenize{{target_outline_str_full}}}, \texttt{\detokenize{{initial_pieces_str}}}, \texttt{\detokenize{{target_str}}}, \texttt{\detokenize{{pieces_str}}}, and \texttt{\detokenize{{few_shot_section}}}, are replaced by the corresponding instance-specific data at inference time.
For multimodal settings, the target image is provided as the visual input together with the textual prompt. The detailed prompt templates for each experimental setting are presented as follows.

\lstdefinestyle{promptstyle}{
    basicstyle=\ttfamily\footnotesize,
    breaklines=true,
    breakatwhitespace=false,
    columns=fullflexible,
    keepspaces=true,
    frame=single,
    backgroundcolor=\color{gray!5},
    xleftmargin=1em,
    xrightmargin=1em,
    aboveskip=0.8em,
    belowskip=0.8em,
    showstringspaces=false
}

% \subsection{Base Prompt}
% \label{app:prompt_base}

\begin{lstlisting}[style=promptstyle,caption={Base prompt template for the standard end-to-end Tangram solution generation setting.},label={lst:prompt_base}]
You are a Tangram Solver Expert. Your task is to solve the puzzle by arranging the 7 given tangram pieces to perfectly fill the target outline shown in the provided image.

Input Data:
1. Target Outline Geometry (Vertices & Edges): {target_outline_str_full}
2. Initial Pieces Specifications: {initial_pieces_str}

Task Requirements:
Determine the final position for each of the 7 pieces.
Important Constraints:
The geometric shape and size of the 7 pieces are FIXED and cannot be modified (rigid transformation only).
The objective is to arrange them so that they completely fill the target area, forming a shape whose outer boundary is identical to the target outline.
There must be no overlaps between pieces and no gaps inside the shape.

Output Format:
Begin by providing your step-by-step spatial reasoning analysis of how you fit the pieces together. Then, at the very end of your response, output the final solution as a single JSON object wrapped in a json code block. The JSON structure must strictly follow this schema:
```json
{ "final_state": { "pieces": [ { "piece_id": "ID", "piece_type": "Type", "transform_matrix": { "translation": {"dx": 0.0, "dy": 0.0}, "rotation_deg": 0.0, "reflection": false }, "vertices": [ {"coords": [x, y], "symbolic": ["val_x", "val_y"]} ] } ] } }
```
\end{lstlisting}

% \subsection{In-Context Learning Prompt}
% \label{app:prompt_icl}

\begin{lstlisting}[style=promptstyle,caption={Prompt template for the in-context learning setting. The solved examples are prepended before the current query while keeping the same task requirements and output schema.},label={lst:prompt_icl}]
You are a Tangram Solver Expert. Your task is to solve the puzzle by arranging the 7 given tangram pieces to perfectly fill the target outline shown in the provided image.

Few-Shot Examples
{few_shot_section}

Current Task
Now, please solve the following specific task based on the image and data provided below:

Current Task Input Data:
1. Target Image Reference: <image>
2. Target Outline Vertices Coordinates and Edges: {target_str}
3. Initial Pieces Specifications: {pieces_str}

Task Requirements:
Determine the final position for each of the 7 pieces.
Important Constraints:
The geometric shape and size of the 7 pieces are fixed and cannot be modified (rigid transformation only).
The objective is to arrange them so that they completely fill the target area, forming a shape whose outer boundary is identical to the target outline.
There must be no overlaps between pieces and no gaps inside the shape.

Output Format:
Begin by providing your step-by-step spatial reasoning analysis of how you fit the pieces together. Then, at the very end of your response, output the final solution as a single JSON object wrapped in a json code block. The JSON structure must strictly follow this schema:
```json
{ "final_state": { "pieces": [ { "piece_id": "ID", "piece_type": "Type", "vertices": [ {"coords": ["latex_x", "latex_y"]} ], "edges": [ {"coords": [["latex_x1", "latex_y1"], ["latex_x2", "latex_y2"]], "length": "latex_len"} ] } ] } }
```
\end{lstlisting}

% \subsection{Visual-Centric Prompt}
% \label{app:prompt_visual_centric}

\begin{lstlisting}[style=promptstyle,caption={Prompt template for the visual-centric setting, where the textual target outline geometry is removed.},label={lst:prompt_visual_centric}]
You are a Tangram Solver Expert. Your task is to solve the puzzle by arranging the 7 given tangram pieces to perfectly fill the target outline shown in the provided image.

Input Data:
1. Please carefully analyze the provided image, where the target outline and relevant vertex coordinates are explicitly displayed.
2. Initial Pieces Specifications: {pieces_str}

Task Requirements:
Determine the final position for each of the 7 pieces.
Important Constraints:
The geometric shape and size of the 7 pieces are fixed and cannot be modified (rigid transformation only).
The objective is to arrange them so that they completely fill the target area, forming a shape whose outer boundary is identical to the target outline.
There must be no overlaps between pieces and no gaps inside the shape.

Output Format:
Begin by providing your step-by-step spatial reasoning analysis of how you fit the pieces together. Then, at the very end of your response, output the final solution as a single JSON object wrapped in a json code block. The JSON structure must strictly follow this schema:
```json
{ "final_state": { "pieces": [ { "piece_id": "ID", "piece_type": "Type", "vertices": [ {"coords": ["latex_x", "latex_y"]} ], "edges": [ {"coords": [["latex_x1", "latex_y1"], ["latex_x2", "latex_y2"]], "length": "latex_len"} ] } ] } }
```
\end{lstlisting}

% \subsection{Chain-of-Thought Prompt}
% \label{app:prompt_cot}

\begin{lstlisting}[style=promptstyle,caption={Prompt template for the chain-of-thought setting, where models are explicitly instructed to reason through silhouette analysis, piece placement, detail filling, and constraint verification before producing the final JSON output.},label={lst:prompt_cot}]
You are a Tangram Solver Expert. Your task is to solve the puzzle by arranging the 7 given tangram pieces to perfectly fill the target outline shown in the provided image.

Input Data:
1. Target Outline Vertices Coordinates and Edges: {target_str}
2. Initial Pieces Specifications: {pieces_str}

Task Requirements:
Determine the final position for each of the 7 pieces.
Important Constraints:
The geometric shape and size of the 7 pieces are fixed and cannot be modified (rigid transformation only).
The objective is to arrange them so that they completely fill the target area, forming a shape whose outer boundary is identical to the target outline.
There must be no overlaps between pieces and no gaps inside the shape.

Chain-of-Thought Reasoning:
Before generating the final coordinate JSON, you must solve this step-by-step:
1. Analyze the Silhouette: Identify key features of the target outline (e.g., right angles, prominent corners, total width/height). Calculate or estimate the bounding box.
2. Strategize Placement: Decide where to place the two Large Triangles first, as they occupy the most space and usually define the main structure.
3. Fill the Details: Determine the positions of the Medium Triangle, Small Triangles, Square, and Parallelogram to fill the remaining gaps. Explain how their edges align with the target outline or other pieces.
4. Verify Constraints: Check if all 7 pieces are used, ensure there are no overlaps, and confirm the total shape matches the target outline coordinates.

Output Format:
Begin by providing your step-by-step spatial reasoning analysis of how you fit the pieces together. Then, at the very end of your response, output the final solution as a single JSON object wrapped in a json code block. The JSON structure must strictly follow this schema:
```json
{ "final_state": { "pieces": [ { "piece_id": "ID", "piece_type": "Type", "vertices": [ {"coords": ["latex_x", "latex_y"]} ], "edges": [ {"coords": [["latex_x1", "latex_y1"], ["latex_x2", "latex_y2"]], "length": "latex_len"} ] } ] } }
```
\end{lstlisting}

% \subsection{Decomposition-Based Prompt}
% \label{app:prompt_decomposition}

\begin{lstlisting}[style=promptstyle,caption={Prompt template for the decomposition-based setting, where models reason using the small isosceles right triangle as the basic geometric unit.},label={lst:prompt_decomposition}]
You are a Tangram Solver Expert. Your task is to solve the puzzle by arranging the 7 given tangram pieces to perfectly fill the target outline shown in the provided image.

Input Data:
1. Target Outline Vertices Coordinates and Edges: {target_str}
2. Initial Pieces Specifications: {pieces_str}

Task Requirements:
Determine the final position for each of the 7 pieces.
Important Constraints:
The geometric shape and size of the 7 pieces are fixed and cannot be modified (rigid transformation only).
The objective is to arrange them so that they completely fill the target area, forming a shape whose outer boundary is identical to the target outline.
There must be no overlaps between pieces and no gaps inside the shape.

Recommended Solving Strategy: Unit Decomposition
Consider solving this puzzle by breaking down the complex shapes into a common geometric basis:
1. Identify the Basic Unit: Interpret the puzzle using a single building block: the small isosceles right triangle (the 'Basic Unit').
2. Quantify the Pieces: Note that every tangram piece is formed by combining these Basic Units:
   - Small Triangle = 1 Unit
   - Square, Medium Triangle, Parallelogram = 2 Units each
   - Large Triangle = 4 Units each
3. Decompose the Target: Mentally overlay a grid of these Basic Units onto the Target Outline. Try to partition the silhouette into regions that match the unit counts of the pieces (e.g., find a 4-unit region for a Large Triangle).
4. Map and Assemble: Assign the pieces to these decomposed regions based on their unit size and shape.

Output Format:
Begin by providing your step-by-step spatial reasoning analysis of how you fit the pieces together. Then, at the very end of your response, output the final solution as a single JSON object wrapped in a json code block. The JSON structure must strictly follow this schema:
```json
{ "final_state": { "pieces": [ { "piece_id": "ID", "piece_type": "Type", "vertices": [ {"coords": ["latex_x", "latex_y"]} ], "edges": [ {"coords": [["latex_x1", "latex_y1"], ["latex_x2", "latex_y2"]], "length": "latex_len"} ] } ] } }
```
\end{lstlisting}

% \subsection{Text-Only Prompt}
% \label{app:prompt_text_only}

\begin{lstlisting}[style=promptstyle,caption={Prompt template for the text-only setting, where the target image is removed.},label={lst:prompt_text_only}]
You are a Tangram Solver Expert. Your task is to solve the puzzle by arranging the 7 given tangram pieces to perfectly fill the specified target outline based on the coordinates provided below.

Input Data:
1. Target Outline Vertices Coordinates and Edges: {target_str}
2. Initial Pieces Specifications: {pieces_str}

Task Requirements:
Determine the final position for each of the 7 pieces.
Important Constraints:
The geometric shape and size of the 7 pieces are fixed and cannot be modified (rigid transformation only).
The objective is to arrange them so that they completely fill the target area, forming a shape whose outer boundary is identical to the target outline.
There must be no overlaps between pieces and no gaps inside the shape.

Output Format:
Begin by providing your step-by-step spatial reasoning analysis of how you fit the pieces together. Then, at the very end of your response, output the final solution as a single JSON object wrapped in a json code block. The JSON structure must strictly follow this schema:
```json
{ "final_state": { "pieces": [ { "piece_id": "ID", "piece_type": "Type", "vertices": [ {"coords": ["latex_x", "latex_y"]} ], "edges": [ {"coords": [["latex_x1", "latex_y1"], ["latex_x2", "latex_y2"]], "length": "latex_len"} ] } ] } }
```
\end{lstlisting}

% \subsection{TCE-Light Prompt}
% \label{app:prompt_tce_light}

\begin{lstlisting}[style=promptstyle,caption={Prompt template for the TCE-light setting, where models output only the rigid transformation parameters rather than full vertices and edges.},label={lst:prompt_tce_light}]
You are a Tangram Solver Expert. Your task is to solve the puzzle by arranging the 7 given tangram pieces to perfectly fill the target outline shown in the provided image.

Input Data:
1. Target Outline Vertices Coordinates and Edges: {target_str}
2. Initial Pieces Specifications: {pieces_str}

Task Requirements:
Determine one rigid transformation for each of the 7 pieces.

Important Constraints:
The geometric shape and size of each piece are fixed and cannot be modified.
Only rigid transformations are allowed: rotation, optional reflection, and translation.
The pieces must completely fill the target area, forming a shape whose outer boundary is identical to the target outline.
There must be no overlaps between pieces and no gaps inside the shape.

Transformation Convention:
For every vertex p=[x,y] of an initial piece, its final position is computed as:
p_final = R(theta) * S(reflection) * p_initial + translation.
Here R(theta) is a counter-clockwise rotation by rotation_degrees.
If reflection is true, S maps [x,y] to [-x,y] before rotation; otherwise S is the identity transform.
translation is [tx, ty] and is added after reflection and rotation.
rotation_degrees should be given in degrees.
translation can be given as decimal numbers or LaTeX-style expressions.

Output Format:
Begin by providing your step-by-step spatial reasoning analysis of how you fit the pieces together. Then, at the very end of your response, output the final solution as a single JSON object wrapped in a json code block.
Do not output vertices or edges. Output only piece_id, rotation_degrees, reflection, and translation.
The JSON structure must strictly follow this schema:
```json
{
  "final_state": {
    "pieces": [
      {
        "piece_id": "ID",
        "rotation_degrees": 0,
        "reflection": false,
        "translation": ["latex_or_decimal_tx", "latex_or_decimal_ty"]
      }
    ]
  }
}
```
\end{lstlisting}

\section{Detailed Experimental Results}
\label{app:detailed_results}
\subsection{Experimental Results for Task 1}
\label{app:task1_results}
The results for Task~1 are summarized in Table~\ref{tab:task1_scores}, revealing substantial performance differences across models. Task~1 serves as a lower-difficulty diagnostic that probes global-shape inference before the more demanding inverse assembly problem in Task~2. Specifically, this task assesses the model's ability to mentally compose a silhouette from disjointed parts and identify the correct visual match accurately. Notably, many models perform poorly on this task, suggesting that Task 1 cannot be reliably solved through simple memorization or superficial pattern matching; instead, it requires genuine compositional spatial reasoning based on part–whole structure. Among open-source baselines, Qwen3-VL-32B-Instruct stands out with an accuracy of 73.05\%, approaching the performance of the closed-source model GPT-5.2, while smaller or earlier architectures struggle to exceed 40\% accuracy. For commercial models, Gemini3-Pro achieves near-ceiling performance with an accuracy of 98.65\%, underscoring its superior capacity for fine-grained visual perception and holistic mental shape composition regarding global silhouettes. For most models, the rate of invalid outputs remains low, suggesting a strong ability to follow instructions and produce interpretable decisions. However, DeepSeek-OCR and DeepSeek-VL2 exhibit noticeably higher invalid rates, suggesting that under the stress of this complex spatial reasoning task, these models often fail to form a decisive conclusion or collapse into generating irrelevant content.

\subsection{Experimental Results for Task 2}
\label{app:task2_results}
Table~\ref{tab:task2_results} presents the quantitative results for Task~2. Although \textbf{Success} is zero for most models, the intermediate metrics reveal why they fail. Most models achieve relatively low \textbf{TSE} scores, typically at the single-digit level, indicating that they can generally produce outputs in the required TCE-style JSON format. This suggests that the main bottleneck is not TCE formatting itself, but compositional spatial reasoning. To avoid over-penalizing minor formatting issues, our evaluation pipeline incorporates a lightweight recovery mechanism: we use \texttt{json\_repair} to fix malformed JSON and regex-based heuristics to resolve common formatting defects in LaTeX-style mathematical expressions. This ensures that we extract and evaluate the model’s true geometric intent.

By contrast, constraint-related errors remain widespread. In particular, \textbf{Rigid Geometry Errors (RGE)} and \textbf{Physical Errors (PE)} are highly prevalent, indicating that models struggle to comply with the fundamental geometric rules of the puzzle. One notable observation is the pronounced discrepancy between silhouette quality and the Success rate. Top-tier models like Claude-Sonnet-4.5 and GPT-5.2 achieve high IoU scores yet fail to produce a single valid solution (0\% Success). Further inspection reveals that the accompanying high RGE and PE scores indicate that this apparent visual fidelity is achieved through ``cheating,'' namely by impermissibly distorting rigid pieces or overlapping them to force a visual match. Meanwhile, Gemini3-Pro stands out as the only model demonstrating robust geometric reasoning, achieving the highest validation pass rate and success score while also maintaining superior silhouette quality.

These intermediate metrics are also useful for tracking incremental progress. Models show substantial variation in \textbf{IoU} and \textbf{Hausdorff Distance}, which remain informative even when \textbf{Success} is zero. Higher IoU corresponds to better global alignment with the target silhouette, while lower Hausdorff Distance captures more accurate boundary matching, providing graded signals below the strict validity threshold. More broadly, these metrics reveal a smoother progression toward successful assembly: improvements typically appear first in \textbf{IoU} (better outline alignment), then in reduced \textbf{RGE} and \textbf{PE} (better adherence to rigid and physical constraints), and finally in non-zero \textbf{Success}. This makes the benchmark suitable for tracking incremental advances, even when achieving full validity remains challenging.

To further validate that these failures reflect model limitations rather than issues in the benchmark construction or evaluation protocol, we also examine the ground-truth TCE solutions. Each TangramPuzzle instance already contains a ground-truth TCE solution. When this ground-truth TCE is evaluated, the seven polygonal pieces are instantiated according to their annotated final states, their union is computed, pairwise overlaps and global connectivity are checked, and the resulting outer boundary is compared with the target outline using the same IoU and Hausdorff Distance metrics as in the main evaluation. The ground-truth TCE achieves 100\% \textbf{VPR} and 100\% \textbf{Success}, confirming that every benchmark instance is constructively solvable and that our automated verifier correctly accepts valid solutions. Therefore, MLLM failures are not caused by invalid target shapes, missing solutions, or an overly strict verifier.

\section{Detailed Analytical Experiments}
\label{app:analysis}

% \subsection{In-context Learning Details}
% \label{In-context Learning Details}
% We examine the effect of In-context Learning (ICL) by prepending three solved examples to the prompt for a subset of models (Qwen3-VL, GPT-5.2, Gemini3-Pro).

% Contrary to the expectation that examples stabilize output, several models exhibited increased Syntax Error (TSE) rates compared to the zero-shot setting. For instance, Qwen3-VL-8B-Instruct and Gemini3-Pro showed higher failure rates in generating valid JSONs. This suggests that the extensive symbolic context imposes a cognitive load that interferes with the model's ability to strictly adhere to structural constraints.

% Despite the syntactic instability, ICL yields clear benefits for silhouette-level quality in valid responses. Models demonstrated improved IoU and reduced Hausdorff distances, effectively learning to better approximate the global target shape. This suggests that while ICL serves as a potent mechanism for refining visual matching strategies and ``teaching'' the task format, it ultimately falls short of spontaneously instilling a fundamental understanding of strict physical validity in models that inherently lack it.

\subsection{In-context Learning Details}
\label{In-context Learning Details}

We examine the effect of In-context Learning (ICL) by prepending three fully solved TCE examples to the prompt for a subset of models, including Qwen3-VL, GPT-5.2, and Gemini3-Pro. The experimental motivation is to disentangle format unfamiliarity from deeper spatial reasoning limitations. Specifically, the few-shot demonstrations explicitly show the required TCE-style output format, including the JSON structure, piece inventory, vertex and edge fields, and coordinate style. Therefore, if the main bottleneck were merely that models were unfamiliar with the TCE schema, providing solved examples should substantially reduce syntax errors and improve the rate of valid assemblies.

However, the results do not support a purely format-unfamiliarity explanation. Originally, Syntax Error (TSE) is already relatively low for most models versus the much higher Rigid Geometry Error (RGE) and Physical Error (PE), suggesting that many models can follow the output format but still fail to satisfy geometric and physical constraints. With ICL, several models even exhibit increased TSE rates compared with the zero-shot setting. For instance, Qwen3-VL-8B-Instruct and Gemini3-Pro more frequently fail to generate valid JSON outputs, indicating that the longer symbolic context can impose an additional cognitive load and destabilize structured generation.

Despite this syntactic instability, ICL sometimes improves silhouette-level quality among successfully parsed responses. Models achieve better IoU or reduced Hausdorff distance in several cases, suggesting that demonstrations can help them better approximate the target shape. Nevertheless, these improvements rarely translate into higher final success, as models frequently violate rigid-body preservation or physical feasibility constraints. 
This implies that few-shot TCE demonstrations are insufficient to induce the compositional spatial reasoning needed to coordinate seven fixed rigid pieces under non-overlap, connectivity, and exact boundary-matching constraints. Thus, the ICL experiment weakens the hypothesis that Task 2 failures stem primarily from unfamiliarity with the TCE schema, instead highlighting constraint-satisfying spatial assembly as the central challenge.

\subsection{Textual Geometry Ablation}
\label{Textual}
To quantify the models' dependency on textual metadata, we evaluated a Visual-Centric setting where explicit textual descriptions of the target outline (vertices and edges) were removed, leaving only the image with annotated coordinates.

The elimination of explicit textual outline specifications significantly destabilized structured generation. Notably, GPT-5.2 experienced a dramatic surge in syntax errors, with TSE rising from $7.19\%$ to $39.67\%$. Similarly, the Deepseek series frequently hallucinated irrelevant content, indicating a heavy reliance on text prompts to guide the generation process.

For the majority of models, the absence of textual data precipitated a marked deterioration in Silhouette Quality, exposing a critical dependency on semantic prompts. This confirms a widespread inability to precisely ``read'' geometric coordinates directly from raw visual inputs. However, Gemini3-Pro's robust performance in this setting serves as a notable exception, highlighting its advanced capability to perform rigorous reasoning solely based on visual grounding, independent of textual crutches.

\subsection{Effect of Chain-of-Thought (CoT)}
\label{CoT}
To investigate whether explicit step-by-step reasoning can alleviate the spatial logic bottleneck, we evaluate models using Chain-of-Thought (CoT) prompting. As shown in the ``Chain-of-Thought (CoT)'' results in Table~\ref{tab:task2_results}, CoT provides only limited and highly uneven benefits. While it improves some intermediate metrics for certain models, it does not fundamentally resolve the core difficulty of the task: producing geometrically valid assemblies under rigid and physical constraints.

We observe that CoT can improve syntax compliance for certain models such as Qwen3-VL-32B and GPT-5.2, suggesting that explicit reasoning prompts may help organize outputs into a more structured form. However, this effect is inconsistent across the board. For example, Gemini3-pro becomes noticeably less stable in its output formatting under CoT, with its TSE rising significantly, indicating that longer reasoning traces can introduce additional formatting noise. This further supports our earlier conclusion that syntax is not the primary bottleneck in Task 2.

More importantly, CoT does not effectively enhance constraint satisfaction. Qwen3-VL-32B managed to achieve a non-zero Success rate ($0.15\%$) under CoT, suggesting that explicit reasoning can occasionally help models stumble upon a valid configuration. Conversely, while silhouette matching slightly improved for GPT-5.2, we observed an exacerbation of geometric deformations, with both PE and RGE increasing. This reinforces our finding that models tend to prioritize "visual cheating" over rigorous physical rules.

For smaller-scale models such as Qwen3-VL-8B, performance remains largely stagnant despite CoT, with Success and VPR rates remaining at zero. This suggests that without basic spatial grounding, adding a thinking step is insufficient to solve complex inverse geometric problems.

\subsection{Effect of Systematic Decomposition}
\label{Decomposition}
Motivated by the human expert analysis in Appendix~\ref{app:human_perf}, we explore whether a decomposition-based strategy can improve performance on Task~2. A key human strategy is to treat both the target silhouette and the seven tangram pieces as compositions of the same basic unit: the small isosceles right triangle. The solver can then reason about how groups of these basic units within the target outline correspond to individual tangram pieces, making the puzzle easier to solve. We therefore introduce prompts that encourage models to perform this kind of decomposition before generating the final assembly.

As shown in the ``Decomposition-Based Strategy'' results in Table~\ref{tab:task2_results}, this strategy leads to modest but consistent improvements in silhouette quality for several models. Qwen3-VL-8B, Qwen3-VL-32B, GPT-5.2, and Gemini3-Pro all achieve higher IoU scores under this setting, suggesting that explicit decomposition helps models align the overall assembly more closely with the target outline. In some cases, boundary precision also improves, suggesting that decomposition provides a more structured basis for global shape planning.

Proprietary models, particularly GPT-5.2 and Gemini3-Pro, show marked improvements in adhering to rigid-body and non-overlap constraints, as reflected by a decrease in RGE and PE scores. Such results underscore the role of a shared geometric basis in guiding models to reconcile how composite shapes populate the available interior space of the target outline. However, Gemini3-Pro's final Success rate experienced a slight decline compared to the original setting. For the Qwen3-VL series, the benefits of decomposition are more subtle; Qwen3-VL-32B showed a marginal improvement in VPR and silhouette quality, while the 8B model's performance remained largely stagnant, with Success and VPR staying at zero. Overall, while encouraging models to reason in terms of a shared geometric basis improves global shape planning and reduces constraint violations for some models, the lack of compositional spatial reasoning remains the primary cause of failure.

\subsection{Effect of Removing Visual Input}
\label{Removing_Visual}
To address the potential concern that the task might degenerate into a purely symbolic encoding--decoding problem once converted into TCE, we conducted a text-only ablation study. In this configuration, models receive only textual descriptions of the vertices of the target outline, with the visual image completely removed. This setup allows us to evaluate whether models can effectively solve the puzzle using symbolic coordinates alone, or whether visual grounding remains indispensable for compositional reasoning.

As illustrated by the ``Text-Only Setting'' results in Table~\ref{tab:task2_results}, removing the visual modality leads to a substantial decline in overall performance. The strongest model, Gemini3-Pro, suffers a substantial drop in both VPR and Success, with both metrics falling by nearly half, while the other models remain unable to produce valid final assemblies. This result indicates that Task~2 is not merely a symbolic coordinate-translation problem: if the task could be solved by directly reading target vertices and converting them into TCE outputs, removing the image should not substantially hurt performance. Instead, the visual silhouette provides a critical global spatial anchor that helps models perceive topology, concavities, boundary structure, and part-to-whole organization, which are difficult to reconstruct from coordinate lists alone.

At the same time, Gemini3-Pro’s sharp performance drop suggests that the model benefits substantially from visual heuristics. This observation is consistent with our broader finding that MLLMs can often produce visually plausible silhouettes while still violating rigid-body or physical constraints. Visual grounding helps the model approximate the global structure, but exact constraint satisfaction still requires reliable symbolic geometric planning, including rigid transformations, non-overlap, connectivity, and full boundary coverage.

The text-only setting does not produce a uniform pattern across intermediate metrics. For some models, certain constraint errors decrease, but these changes do not translate into better final outcomes. In other words, access to coordinates alone may help with parts of the formal representation, yet it remains insufficient for carrying out the full part-to-whole assembly process. The persistent gap in Success indicates that solving the task requires more than simply decoding symbolic input into TCE format.

\subsection{Effect of TCE-Light Representation}
\label{app: TCE-Light}
To examine whether Task~2 failures are caused by the complexity of the full TCE representation rather than spatial reasoning itself, we introduce a simplified \textit{TCE-Light} setting. Instead of requiring models to generate complete vertex and edge lists for every piece, TCE-Light asks them to output only one rigid transformation for each piece, including \texttt{piece\_id}, rotation angle, optional reflection, and translation. The final vertices are then deterministically computed by applying the predicted transformations to the initial pieces. This design substantially reduces the burden of symbolic coordinate generation and enforces rigid-shape preservation by construction, making RGE zero for all successfully parsed outputs.

As shown in Table~\ref{tab:task2_results}, however, simplifying the output format does not solve the task. Although TCE-Light removes rigid-geometry errors, models still exhibit high PE, low VPR, and near-zero Success. For example, Qwen3-VL-32B-Instruct and GPT-5.2 achieve 0.00\% Success under TCE-Light, while Gemini3-Pro's success rate drops to 3.59\%. Moreover, silhouette quality often drops substantially, with lower IoU and larger Hausdorff distance compared with the full TCE setting. These findings suggest that the primary barrier is not the ability to write detailed TCE coordinates. Even when the output is reduced to simple rigid transformations, MLLMs still struggle to determine physically feasible placements that jointly satisfy non-overlap, connectivity, and target-boundary constraints.

\subsection{Relationship between Task 1 and Task 2}
\label{app:task_relation}
Task~1 and Task~2 are designed to probe complementary levels of compositional spatial reasoning. Serving as a foundational diagnostic, Task~1 uses a multiple-choice format for lightweight inference of part-to-whole global shapes, which enables clean and objective automatic evaluation. In contrast, Task~2, our primary constructive evaluation, requires models to independently generate a complete assembly under strict rigid-body, non-overlap, connectivity, and boundary-matching constraints.

To better understand the relationship between the two tasks, we compute cross-model correlations between Task~1 Accuracy and the main Task~2 metrics using the models evaluated on both tasks. We report both Pearson and Spearman correlations, where Pearson correlation measures linear association between the scores while Spearman correlation assesses ranking consistency across tasks. As shown in Table~\ref{tab:task_correlation}, Task~1 Accuracy correlates strongly with Task~2 IoU, suggesting that Task~1 captures a capability that is useful for Task~2: composing local pieces into a plausible global silhouette. Models that perform better on Task~1 tend to achieve better silhouette overlap in Task~2.

However, Task~1 does not reliably predict whether a model can produce a fully valid final assembly. Although Pearson correlations with VPR and Success are moderately positive, the corresponding Spearman correlations are much weaker, indicating that the relationship is not robust in rank order and is affected by the sparsity of successful Task~2 solutions. For example, models such as GPT-5.2 and Qwen3-VL-32B-Instruct achieve strong Task~1 accuracy but still obtain 0\% Success in Task~2 under the standard setting. This indicates that Task~2 requires additional capabilities beyond global-shape recognition, including preserving rigid piece geometry, avoiding overlaps, maintaining connectivity, and exactly matching the target boundary. Overall, the two tasks form a hierarchical evaluation: Task~1 measures the perceptual prerequisite of part-to-whole silhouette composition, while Task~2 measures inverse generative assembly under strict geometric constraints.

\begin{table}[t]
\centering
% \small
% \renewcommand{\arraystretch}{1.15}
% \setlength{\tabcolsep}{9pt}
\caption{Cross-model correlation between Task~1 Accuracy and Task~2 metrics.}
\label{tab:task_correlation}
\resizebox{\linewidth}{!}{
\begin{tabular}{lcccccc}
\toprule
\textbf{Correlation} & \textbf{RGE} & \textbf{PE} & \textbf{VPR} & \textbf{IoU} & \textbf{Hausdorff} & \textbf{Success} \\
\midrule
Pearson  & -0.36 & -0.51 & 0.69 & 0.82 & -0.53 & 0.69 \\
Spearman & -0.57 & -0.32 & 0.30 & 0.79 & -0.68 & 0.27 \\
\bottomrule
\end{tabular}
}
\end{table}

\section{Detailed Human Performance Analysis}
\label{app:human_perf}

\subsection{Human Performance on Task~1}
\label{app:human_perf1}
We evaluate human performance on Task~1 using three domain experts, each of whom completed a random subset of 50 questions. As shown in Table~\ref{tab:human_performance}, all three participants achieved 100\% accuracy. This indicates that Task~1 is straightforward for humans: once the seven tangram pieces are composed, the correct outer silhouette is typically easy to recognize. Participants also reported low complexity and short time usage, suggesting that the task can be solved efficiently and with little difficulty by human experts.

The human results also help contextualize the model performance in Table~\ref{tab:task1_scores}. Although Gemini3-Pro achieves near-ceiling accuracy and GPT-5.2 also performs strongly, most other models remain substantially below the human level, with many open-source models still far from reliable recognition. In particular, model accuracies span a wide range, from below 30\% to nearly 99\%, revealing a clear gap between proprietary and many open-source models. This spread indicates that Task~1 is far from saturated: despite being easy for humans, it remains discriminative for models and provides a meaningful test of visual perception and mental shape composition.

\subsection{Human Performance on Task~2}
\label{app:human_perf2}

We evaluate human performance on Task~2 by involving three domain experts who independently solved the task sets under the same conditions as the models.

\paragraph{Performance Patterns.}
As shown in our results, human solvers rarely violate fundamental geometric constraints such as piece shape preservation or piece count correctness. Consequently, human performance manifests as a binary outcome: a solution is either strictly valid or the participant fails to find a solution within a reasonable time. This stands in sharp contrast to model predictions, which often achieve partial visual similarity through impermissible physical violations (e.g., overlapping pieces or shape distortion).

\paragraph{Solving Strategies.}
\label{Solving Strategies}
% We observe substantial variance in success rates linked to the strategies employed. One expert achieved perfect performance by adopting a \textit{systematic decomposition strategy}: all tangram pieces were interpreted as combinations of congruent isosceles right triangles (with unit legs and hypotenuse $\sqrt{2}$). By decomposing both the target silhouette and the pieces into this common geometric basis, the solver could explicitly reason about how composite shapes fit within the outline. In contrast, the other two experts primarily relied on \textit{iterative trial-and-error}, repeatedly adjusting piece positions until a feasible configuration emerged. This heuristic approach resulted in lower success rates and significantly longer solving times.

We observe substantial variance in success rates across the strategies employed. One expert achieved perfect performance by adopting a \textit{systematic decomposition strategy}. Specifically, all seven tangram pieces were interpreted as compositions of a common geometric primitive: the small isosceles right triangle. Under this view, each piece corresponds to a fixed number and arrangement of these primitive units, and the target silhouette can likewise be mentally decomposed into the same unit-triangle basis. Once the silhouette is represented in this way, the puzzle becomes a structured assignment problem: the solver searches for regions within the decomposed outline whose unit counts and geometric patterns match those of the seven tangram pieces, and then assembles the solution by mapping pieces to these regions. In contrast, the other two experts primarily relied on \textit{iterative trial-and-error}, repeatedly adjusting piece positions until a feasible configuration emerged. This heuristic approach resulted in lower success rates and significantly longer solving times.

\paragraph{Complexity Assessment.}
Participants were asked to rate task difficulty on a scale from 1 to 5. The high average complexity score of 4.1 confirms that TangramPuzzle is non-trivial even for humans. On average, each instance required approximately 6.7 minutes to solve, reflecting the substantial cognitive effort required for precise geometric reasoning and assembly planning.

\section{Human Evaluation on MLLMs}
\label{app:Human_Evaluation}
% 人类来评估模型做得好坏，这里需要设定一些让人来评估的指标，从现有的自动指标评估不到的维度进行评估
% Automatic metrics primarily capture strict geometric validity, often overlooking perceptual plausibility. To address this, we conduct a human evaluation on Task~2 using three criteria rated on a 5-point scale: Perceptual Silhouette Match (PSM), which assesses the visual similarity of the assembly to the target; Piece Plausibility \& Canonicality (PPC), which evaluates whether piece placements appear natural rather than visually strained; and Near-Miss Severity (NMS), which measures the proximity of an incorrect prediction to a successful solution. Table~\ref{tab:Human evaluation} summarizes the human evaluation results. Gemini3-Pro consistently receives high scores, indicating strong perceptual alignment with the target silhouette; even when automatic success is not achieved, many solutions are judged to be close to correct. In contrast, GPT-5.2 receives substantially lower scores, with frequent perceptual mismatches, implausible constructions, and failures that are judged to be far from correct.

Automatic metrics primarily capture strict geometric validity, often overlooking perceptual plausibility. To address this, we conduct a human evaluation on Task~2 using three criteria rated on a 5-point scale: Perceptual Silhouette Match (PSM), which assesses the visual similarity of the assembly to the target; Piece Plausibility \& Canonicality (PPC), which evaluates whether piece placements appear natural rather than visually strained; and Near-Miss Severity (NMS), which measures the proximity of an incorrect prediction to a successful solution. Table~\ref{tab:Human evaluation} summarizes the human evaluation results, where the average scores are reported as mean $\pm$ sample standard deviation across the three experts. Gemini3-Pro consistently receives high scores, indicating strong perceptual alignment with the target silhouette; even when automatic success is not achieved, many solutions are judged to be close to correct. In contrast, GPT-5.2 receives substantially lower scores, with frequent perceptual mismatches, implausible constructions, and failures that are judged to be far from correct. The relatively low inter-rater variance, ranging from 0.15 to 0.36 across all reported criteria, further supports the reliability and consistency of the human evaluation process.

\begin{table}[t]
\centering
\caption{Human evaluation of MLLM on Task~2.}
\label{tab:Human evaluation}
\resizebox{\linewidth}{!}{
\begin{tabular}{l l c c c}
\toprule
\textbf{Model} & \textbf{Expert} & \textbf{PSM} ($\uparrow$) & \textbf{PPC} ($\uparrow$) & \textbf{NMS} ($\uparrow$) \\
\midrule
\multirow{4}{*}{Gemini3-Pro}
 & Expert 1 & 04.20   & 04.30 & 04.60 \\
 & Expert 2 & 04.50   & 04.60   & 04.10   \\
 & Expert 3 & 04.20 & 04.50 & 04.00 \\
 & Average   & $\bm{04.30 \pm 0.17}$   & $\bm{04.47 \pm 0.15}$   & $\bm{04.23 \pm 0.32}$   \\
\midrule
\multirow{4}{*}{GPT-5.2}
 & Expert 1 & 01.60   & 01.50 & 01.80 \\
 & Expert 2 & 01.20   & 01.10   & 01.10   \\
 & Expert 3 & 01.50 & 01.20 & 01.60 \\
 & Average   & $\underline{01.43 \pm 0.21}$   & $\underline{01.27 \pm 0.21}$   & $\underline{01.50 \pm 0.36}$   \\
\bottomrule
\end{tabular}
}
\end{table}

\section{Detailed Case Study}
\label{app:Case_Study}
\subsection{Visual Examples}
Figure~\ref{fig:results} visualizes the generation results for a U-shaped target (Task 2). Gemini3-Pro achieves a perfect assembly, demonstrating its ability to effectively ground coordinate constraints to solve complex compositional problems. A critical failure mode observed across other models is the tendency to prioritize visual silhouette matching at the expense of geometric constraints. For instance, Claude-Sonnet-4.5 attempts to force an alignment with the target contour by non-rigidly elongating the edges of the left and right triangles, whereas InternVL3-78B generates severe overlaps to maximize area coverage. Meanwhile, GPT-5.2 produces a visually plausible silhouette but fails to preserve component fidelity, utilizing an incorrect piece inventory (e.g., hallucinating an extra square while omitting the parallelogram). We further observe that smaller models such as Qwen3-VL-8B-Instruct often arrange pieces in a simple linear sequence, indicating an inability to map symbolic coordinate specifications onto a 2D planar workspace.

\subsection{Fine-grained Insights into Physical Errors}
\label{sec:fine_grained_physical_errors}

To further understand why models fail to produce physically valid tangram assemblies, we decompose the original Physical Error (PE) metric into two independent sub-metrics: \textbf{Overlap Error (OE)} and \textbf{Connectivity Error (CE)}. OE measures the rate of predictions containing impermissible area overlap between any pair of pieces, while CE measures the rate of predictions whose piece union fails to form a single connected component. The original PE metric is retained as an aggregate physical-validity indicator:
\begin{equation}
    \mathrm{PE} = \mathrm{OE} \lor \mathrm{CE}.
\end{equation}

Table~\ref{tab:physical_error_breakdown} reports the fine-grained breakdown. The results reveal that connectivity is a particularly challenging constraint for current MLLMs. Across most models, CE is higher than OE, indicating that models are not only prone to local collision errors, but are even more likely to scatter pieces across the canvas or form multiple disconnected groups instead of assembling a single cohesive tangram object. For example, Gemini3-Pro achieves the lowest OE among all evaluated models, with only $29.19\%$ overlap errors, but its CE remains high at $62.87\%$. This suggests that even when a model can partially avoid local overlaps, it still struggles to enforce global connectedness.

The decomposition also shows that physical invalidity often arises from multiple interacting failure modes. For several models, both OE and CE are high, meaning that their predictions simultaneously contain local overlaps and global disconnections. This supports our broader finding that current MLLMs struggle not only with local geometric placement, but also with global compositional organization under strict physical constraints.

\begin{table}[t]
\centering
\small
\caption{
Fine-grained breakdown of Physical Errors.
}
\resizebox{\linewidth}{!}{
\begin{tabular}{lccc}
\toprule
\textbf{Model} & \textbf{OE $\downarrow$} & \textbf{CE $\downarrow$} & \textbf{PE $\downarrow$} \\
\midrule
Qwen3-VL-8B~\cite{bai2025qwen3vltechnicalreport}  & \underline{36.23} & \underline{79.19} & \underline{80.69} \\
Qwen3-VL-32B~\cite{bai2025qwen3vltechnicalreport} & 79.64 & 88.17 & 94.46 \\
InternVL3-78B~\cite{zhu2025internvl3}         & 86.38 & 85.63 & 92.96 \\
GLM-4.6V~\cite{zai2025glm46v}             & 81.59 & 88.62 & 93.56 \\
GPT-5.2~\cite{openai2025gpt52}               & 81.74 & 82.93 & 92.81 \\
Gemini3-Pro~\cite{deepmind2025gemini3pro}           & \textbf{29.19} & \textbf{62.87} & \textbf{65.57} \\
Claude-Sonnet-4.5~\cite{anthropic2025claudesonnet45}     & 70.21 & 90.72 & 97.90 \\
\bottomrule
\end{tabular}
}
\label{tab:physical_error_breakdown}
\end{table}

\section{Discussion}
\label{sec:discussion}
\subsection{Generalizability to Real-World Spatial Scenarios}
Although TangramPuzzle is instantiated in a highly controlled puzzle domain, the benchmark targets a set of core capabilities that extend beyond tangrams themselves and are broadly relevant to real-world spatial reasoning. Fundamentally, solving a tangram requires preserving the identity of rigid parts under Euclidean transformations, composing multiple local pieces into a globally valid configuration, and satisfying hard geometric constraints such as non-overlap, connectivity, and exact boundary alignment. These requirements mirror the structural demands of practical spatial tasks, where success depends not only on recognizing a plausible overall shape, but also on maintaining the validity of each constituent part throughout construction.

This is particularly relevant to applications such as CAD-style assembly and robotic manipulation. In CAD assembly, parts have fixed geometry and must be placed through rigid transformations while satisfying contact, fit, and compatibility constraints; a visually plausible arrangement is insufficient if dimensions, orientations, or interfaces are inconsistent. The Tangram Construction Expression (TCE) explicitly reflects the parametric representations used in CAD software. Similarly, robotic manipulation requires translating visual goals into precise, collision-free spatial plans for rotating, translating, and composing objects into a target structure. Our deterministic evaluation engine mirrors these requirements through programmatic checks for geometric validity. TangramPuzzle abstracts them into a minimal 2D setting, enabling evaluation of whether a model can preserve rigid-body structure and compose parts coherently, rather than merely relying on coarse silhouette matching. These capabilities are essential for integrating AI into reliable design and assembly workflows.

The controlled nature of the benchmark is therefore a deliberate design choice. Real-world spatial tasks typically involve additional sources of difficulty, including perception noise, occlusion, uncertain correspondences, and underspecified goals. While these factors are important, they can also obscure the underlying reasoning problem and make evaluation ambiguous. By abstracting away such confounds, TangramPuzzle isolates the compositional geometric core of the problem and enables exact, machine-verifiable assessment. This is aligned with the motivation of our symbolic TCE representation and constraint-based verifier, which were introduced precisely to move evaluation beyond approximate perceptual similarity toward rigorous geometric correctness. In this sense, the benchmark should be viewed as a diagnostic testbed for compositional spatial reasoning abilities that real-world systems also rely upon.

\subsection{Unique Challenges of TangramPuzzle}
TangramPuzzle poses a form of spatial reasoning that is structurally different from many existing benchmarks. Rather than asking the model to recognize a local spatial relation, answer a geometry-related question, or reason over an already observed scene, it requires the model to construct a globally valid configuration from multiple rigid parts under strict constraints.

\paragraph{Diagnostic interpretability.} 
One advantage of TangramPuzzle is that it does not collapse evaluation into a single accuracy or success score. Because the benchmark separates failures into syntax errors, rigid-geometry violations, physical-validity violations, and silhouette mismatch, it reveals not only whether a model fails, but also \emph{how} it fails. This makes the benchmark more informative for diagnosing weaknesses and tracking progress.

\paragraph{Rigid-body preservation.} A distinctive challenge of TangramPuzzle is that every piece is a rigid geometric object whose shape must remain unchanged. Our experiments reveal a systematic failure mode in which MLLMs appear to “cheat” by implicitly deforming pieces to better match the target silhouette. This type of error is easy to miss when evaluation relies mainly on answer correctness or approximate outline similarity, but TangramPuzzle makes it explicit by checking whether each piece preserves its exact geometry.

\paragraph{Inverse assembly.} TangramPuzzle is fundamentally an inverse assembly problem. The model must search over a large transformation space—including translations, rotations, and reflections—to find a configuration that satisfies both the outer boundary and the internal non-overlap constraints. This is more demanding than simply interpreting a given spatial scene, because the model must reason from the desired global shape back to the placement of individual parts.

\paragraph{Mathematically grounded validity.} 
% Rather than evaluating spatial reasoning through natural-language answers or coarse perceptual judgments, TangramPuzzle grounds the task in an explicit geometric representation via TCE, where each solution is specified by exact coordinates and structural relations. This makes the reasoning process mathematically explicit and the final output exactly verifiable. As a result, the benchmark can sharply distinguish between outputs that merely \emph{look} correct and those that are truly geometrically valid: a predicted assembly may closely match the target silhouette while still violating rigidity or non-overlap constraints, and is therefore marked invalid.

Rather than evaluating spatial reasoning through natural-language answers or coarse perceptual judgments, TangramPuzzle grounds the task in explicit geometry via TCE, where each solution is specified by exact coordinates and structural relations, making the output exactly verifiable. As a result, the benchmark can distinguish outputs that merely \emph{look} correct from those that are truly geometrically valid: a predicted assembly may closely match the target silhouette while still violating rigidity or non-overlap constraints, and thus be invalid.

\subsection{Future Work}
While the current benchmark utilizes the classic seven-piece set, the TCE framework inherently represents pieces by types, vertices, edges, and transformations. This representation can be directly extended to include more than seven polygonal components, variable piece inventories, or irregular dissection puzzles. The existing TCE schema can seamlessly record these new shapes, maintaining a consistent evaluation protocol while significantly increasing combinatorial difficulty.

A direction for future work is to bridge the current controlled setting with more realistic spatial reasoning scenarios. The coordinate-based TCE can be naturally upgraded to a 3D variant, where 2D polygons become polyhedra or voxel-based components, and transformations expand to 6-Degrees-of-Freedom (3D translations and rotations). Consequently, the verifier would shift from 2D polygon intersection to 3D collision detection, volumetric coverage, and surface alignment checks. This progression would allow TangramPuzzle-style evaluation to extend to polycube or CAD-style assembly tasks, while preserving the core principle of exact, machine-verifiable geometry.

To improve model performance on TangramPuzzle, future work could explore inference-time reasoning systems that couple generation with deterministic verification. In particular, test-time computation methods such as verifier-guided search, Monte Carlo Tree Search, and iterative self-refinement could exploit the deterministic TCE verifier to explore and repair candidate assemblies. Code-Interpreter-based verification loops could allow models to iteratively validate and correct physical overlap errors after generating an initial TCE. Multi-agent formulations may also be effective: different agents could specialize in proposing placements, checking topological overlaps, verifying connectivity, evaluating global silhouette matching, and revising invalid configurations.

\end{document}